\documentclass[acmsmall,screen]{acmart}

\AtBeginDocument{%
  }

\setcopyright{cc}
\setcctype{by}
\acmDOI{10.1145/3798237}
\acmYear{2026}
\acmJournal{PACMPL}
\acmVolume{10}
\acmNumber{OOPSLA1}
\acmArticle{129}
\acmMonth{4}
\received{2025-10-10}
\received[accepted]{2026-02-17}

\usepackage[T1]{fontenc}

\usepackage{amsmath,amssymb,amsfonts}
\usepackage{bm}
\usepackage[utf8]{inputenc}
\usepackage{cases}
\usepackage{enumitem}

\usepackage{booktabs}
\usepackage{multirow}
\usepackage{makecell}
\usepackage{tabularx}

\usepackage[ruled,vlined]{algorithm2e}

\usepackage{graphicx}
\usepackage{xspace}
\usepackage{hhline}
\usepackage{hyperref}

\usepackage{subcaption}

\usepackage{wrapfig}  

\usepackage[nameinlink,capitalize,noabbrev]{cleveref}
\usepackage{longtable}
\usepackage{adjustbox}

\begin{document}


\title{SART: Sign-Absolute Reformulation Theory for Binary Variable Reduction in Neural Network Verification}


\author{Jin Xu}
\email{jin.xu@tongji.edu.cn}
\orcid{0009-0000-3791-6336}
\affiliation{%
  \institution{Tongji University}
  \city{Shanghai}
  \country{China}
}

\author{Miaomiao Zhang}
\affiliation{%
  \institution{Tongji University}
  \city{Shanghai}
  \country{China}}
\email{miaomiao@tongji.edu.cn}
\orcid{0000-0001-9179-0893}
\authornote{Corresponding authors.}

\author{Bowen Du}
\affiliation{%
  \institution{Tongji University}
  \city{Shanghai}
  \country{China}}
\email{bowendu@tongji.edu.cn}
\orcid{0000-0002-3755-4870}
\authornotemark[1]


\begin{abstract}

Complete formal verification of neural networks is crucial for their deployment in safety-critical domains. A key bottleneck stems from encoding complexity: traditional methods assign one binary variable per unstable ReLU neuron. We propose the Sign-Absolute Reformulation Theory (SART), which fundamentally breaks the conventional one-to-one mapping between unstable neurons and binary variables by establishing formal reducibility criteria. This allows for finer-grained modeling, where each unstable neuron corresponds on average to fewer than one binary variable, thereby reducing verification complexity at its source.
Based on SART, we derive a  theoretical lower bound on the number of binary variables required for complete verification and, under the assumption that $P \neq NP$, prove that variables in the final layer can be compressed by 50\%, while the number of variables in intermediate layers cannot be further reduced. 
To overcome the apparent ``last-layer-only'' limitation, we recast verification as a sequential process and, crucially, show that the gain lifts to the entire network: LayerABS, a SART-based progressive tightening verifier, iteratively treats intermediate layers as temporary final layers and propagates tight bounds that shrink the global search space and binary-variable counts.
Furthermore, we reveal a structural law influencing verification complexity: when the signs of weights of unstable neurons satisfy numerical symmetry, with positive and negative weights equal or differing by at most one, the worst-case verification complexity achieves the theoretical optimum, offering theoretical guidance for the design of verification-friendly architectures.
As a general-purpose underlying encoding, the value of SART is independent of specific algorithms. To comprehensively evaluate its effectiveness, we first evaluate the abstraction-free SART encoding, and then integrate it with abstraction techniques to construct the complete verifier LayerABS and its incomplete variant Incomplete-LayerABS. 
Across benchmarks, our methods surpass state-of-the-art baselines, validating SART's practical impact.

\end{abstract}




\maketitle

\section{Introduction}                                                                                          
Neural networks have significant applications in many safety-critical domains, particularly in aerospace, autonomous driving, and other fields requiring high reliability. The emergence of adversarial examples~\cite{goodfellow2014explaining} demonstrates that neural networks may produce incorrect outputs in response to minor input perturbations. Therefore, it is necessary to verify neural networks to ensure their robustness.   
In neural network verification, unstable neurons refer to ReLU neurons whose activation states cannot be determined within a given input range. The output interval of a ReLU function must be computed, considering cases where the input may take both positive and negative values.  
Complete verification requires accounting for both possible activation paths of each unstable neuron, and typically achieves this by reformulating the verification problem using one of three paradigms: SMT-based methods \cite{biere2009conflict}, MILP-based methods \cite{floudas2005mixed}, or BaB-based methods \cite{lawler1966branch}.


\begin{figure}[htbp]
  \centering
  \begin{subfigure}[t]{0.35\textwidth}
    \includegraphics[width=\textwidth]{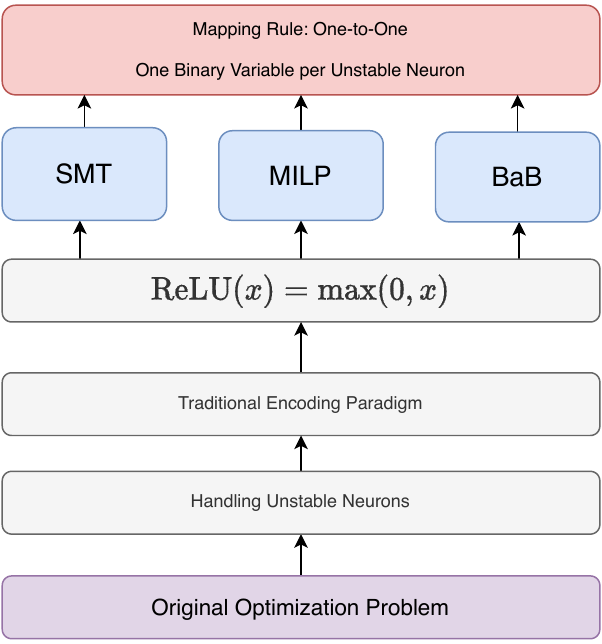}
    \caption{Traditional Methods}
  \end{subfigure}
  \hspace{0.03\textwidth}  
  \begin{subfigure}[t]{0.46\textwidth}
    \includegraphics[width=\textwidth]{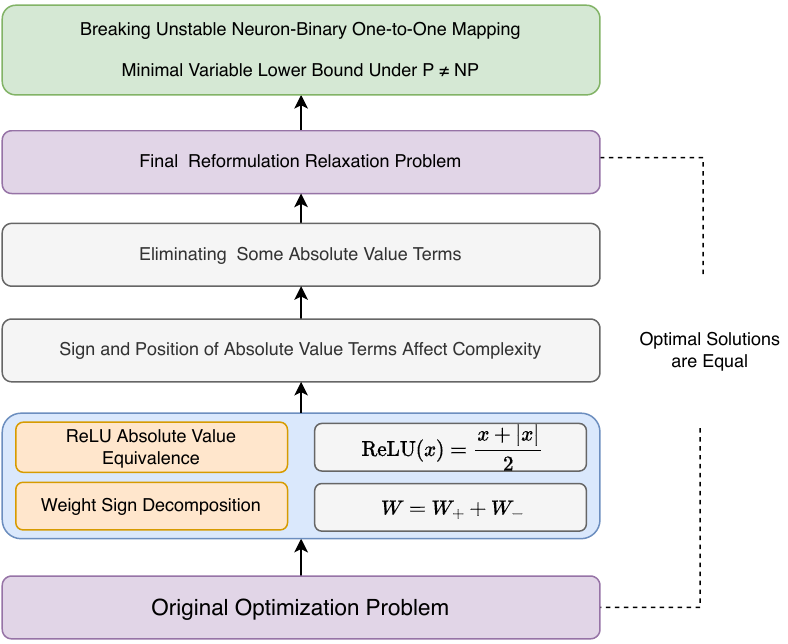}
    \caption{\makebox[0.32\linewidth][l]{SART}}
  \end{subfigure}
  \caption{The evolution of complete verification. \textbf{a:} The traditional method produces binary variables equal to the number of unstable neurons. \textbf{b:} Our sign-absolute reformulation theory generates fewer binary variables.}
\end{figure}


For the SMT-based category, representative algorithms include Reluplex ~\cite{katz2017reluplex}, which first proved that neural network verification is an NP-complete problem. Marabou~\cite{katz2019marabou, wu2024marabou} improves upon Reluplex. The NP-completeness of SMT problems implies that worst-case performance must be anticipated. In these state-of-the-art SMT verification algorithms, stable neurons (activated or non-activated) are not split, whereas unstable neurons must be encoded with an additional binary variable.  
For the MILP-based category, neural networks are primarily modeled using Big-M formulations \cite{fischetti2018deep,cheng2017maximum,anderson2020strong}. These formulations introduce one binary variable for each unstable neuron. The most advanced MILP verification algorithm, MIPVerify \cite{tjeng2017evaluating}, improves upon traditional MILP methods by employing novel preprocessing techniques that reduce the search space by one or more orders of magnitude. However, it still requires one binary variable per unstable neuron during encoding.  
For the BaB-based category, representative algorithms include BaB \cite{bunel2018unified} and its extensions \cite{bunel2020branch,lu2019neural}. Some approaches leverage GPU acceleration for parallelized BaB solving, such as $\beta$-CROWN \cite{wang2021beta} and MNBAB \cite{ferrari2022complete}. These verification algorithms rely on branch-and-bound techniques, recursively splitting unstable neurons. Each node generates two branches, estimating the lower and upper bounds of the output. If the lower bound of a branch exceeds the current global optimum's upper bound, the branch is pruned. Otherwise, key neurons are prioritized for splitting, iteratively updating the global solution bounds. Essentially, BaB-based algorithms also introduce one binary variable for each unstable neuron.  
As seen across all three problem-solving approaches, each unstable neuron requires a corresponding binary variable. This one-to-one mapping between unstable neurons and binary variables is referred to as the neuron-variable rigid mapping.  
As the number of binary variables increases, it leads to a combinatorial explosion problem. Therefore, in neural network verification modeling, the number of binary variables corresponding to unstable neurons is the most critical factor restricting the efficiency of complete solving.

This study is the first to fundamentally reveal that unstable ReLU neurons do not have a one-to-one correspondence with binary variables. By introducing weight sign decomposition and the equivalent transformation of the ReLU activation into absolute value functions, we propose the Sign-Absolute Reformulation Theory (SART), breaking the traditional one-to-one neuron-variable mapping paradigm.
Existing theory~\cite{xu1987special} shows that in minimization problems, when the coefficients of absolute value terms in the objective are positive, the original mixed-integer program degenerates to a linear program. Building on this, we further analyze the case of negative coefficients, demonstrating that the original complexity is retained and revealing the essential impact of the sign of absolute value coefficients on problem complexity. This conclusion extends to constraints involving absolute value terms, proving that binary variables are required regardless of coefficient signs, thus highlighting the critical impact of their occurrence in either the objective function or the constraints on problem scale. 
Based on these results, we systematically summarize the decisive roles of the sign and position (objective or constraint) of absolute value terms on the number of binary variables and derive linearization criteria for non-nested absolute values. This provides a theoretical foundation for subsequent reducibility criteria of unstable neurons.

Based on this theory, we propose the SART encoding method for neural network verification. This enables finer-grained modeling, where each unstable neuron corresponds to fewer than one binary variable on average.
In interval-complete bound computation, SART proves that the total number of binary variables reaches the theoretical lower bound: only the final layer can be simplified, halving the binary variables for unstable neurons; intermediate layers cannot be further compressed, with each unstable neuron still requiring one binary variable. Under the assumption \( P \neq NP \)~\cite{baker1975relativizations}, SART establishes the theoretical minimum number of binary variables under completeness conditions, where any further compression would violate complexity theory or cause loss of completeness.

From a static perspective, this may look like bad news for verification: if the entire task is treated as a single monolithic MILP, compressing only the final layer seems of limited value. However, the bad news carries good news. Verification is inherently a dynamic, sequential process of information accumulation and can be viewed as a series of tightly coupled small MILPs, where earlier layers directly influence later ones and the bottleneck lies in the search space. Building on this view, we propose a SART-based progressive tightening strategy,  systematically realized as the LayerABS algorithm. Crucially, we show that the apparent ``last-layer-only'' gain lifts to the entire network by recasting verification as a sequential process: LayerABS iteratively treats intermediate layers as temporary final layers and applies SART, achieving about \(50\%\) local variable reduction per step; the resulting high-precision bounds propagate downstream, fixing many unstable neurons in subsequent layers early and markedly reducing the global number of binary variables. Through this chain effect in dynamic verification, SART yields network-wide acceleration.
Further theoretical analysis shows that for networks with mixed-sign weights, when the sign distribution of weights associated with unstable neurons is numerically symmetric, where the numbers of positive and negative weights are equal or differ by at most one, one attains optimal worst-case verification complexity, offering theoretical guidance for the design of verification-friendly architectures.

As a general low-level encoding framework, SART is valuable independently of any specific algorithm. 
To comprehensively assess its performance, we build a multi-level evaluation suite: we first evaluate the abstraction-free SART encoding (PureSART); then, combining it with abstraction, we construct the complete verifier LayerABS and its incomplete variant Incomplete-LayerABS. Experiments show that these SART-based instances significantly outperform related state-of-the-art methods across multiple benchmarks.
Specifically, compared with Standard MILP, PureSART improves the success rate while reducing verification time by up to 34.4\%; the complete verifier LayerABS lowers timeout rates by up to 22\%, 53\%, and 40\% relative to MIPVerify, Marabou, and $\beta$-CROWN, respectively, and reduces verification time by up to 58.3\%, 74.5\%, and 66.7\%; the incomplete variant Incomplete-LayerABS, compared with PRIMA~\cite{muller2022prima}, achieves a higher success rate while further shortening verification time by 44.6\%. 
\textbf{The core contributions are as follows}:
\begin{itemize}
    \item Proposing the SART encoding theory for neural network verification, achieving fewer than one binary variable per unstable neuron on average. Under interval-complete bound computation, SART reduces final-layer variables by 50\% and proves that intermediate layers are incompressible, establishing the theoretical minimal variable count under \( P \neq NP \).
    \item We overcome the apparent ``last-layer-only'' limitation by recasting verification as a sequential process and, crucially, show that the gain lifts to the entire network: LayerABS implements a SART-based progressive tightening that treats intermediate layers as temporary final layers, achieving about \(50\%\) local variable cuts per step and propagating tight bounds that shrink the global search space and binary-variable counts.
    \item Revealing the role of sign symmetry in the weights of unstable neurons in enabling the attainment of theoretical optimal worst-case complexity, thereby offering theoretical guidance for the design of verification-friendly architectures.
    \item We first evaluate the abstraction-free SART encoding (PureSART), and then combine SART with abstraction techniques to construct the complete verifier LayerABS and the incomplete variant Incomplete-LayerABS. These SART-based algorithms significantly outperform related SOTA methods across multiple benchmarks, demonstrating the effectiveness of both the encoding and the algorithms.
\end{itemize}

\section{Preliminaries}
In this section, we provide the definition of feedforward ReLU neural networks, the optimization formulation of neural network verification, ReLU activation state classification, and binary variable transformation.

\subsection{Feedforward ReLU Neural Network}
\label{sec:architecture}

An \( L \)-layer feedforward neural network \( f(\mathbf{x}): \mathbb{R}^d \rightarrow \mathbb{R}^m \) is defined as follows. The input layer (\(\ell = 0\)) receives \(\mathbf{x} \in \mathbb{R}^d\) with activation \(\mathbf{a}^{(0)} = \mathbf{x}\). For hidden layers (\(\ell = 1,\ldots,L-1\)), the pre-activation and activation are computed as:
\begin{align}
  \mathbf{z}^{(\ell)} &= \mathbf{W}^{(\ell)}\mathbf{a}^{(\ell-1)} + \mathbf{b}^{(\ell)}, \quad \mathbf{W}^{(\ell)} \in \mathbb{R}^{n_\ell \times n_{\ell-1}}, \ \mathbf{b}^{(\ell)} \in \mathbb{R}^{n_\ell}, \\
  \mathbf{a}^{(\ell)} &= \mathrm{ReLU}(\mathbf{z}^{(\ell)}) = \max(\mathbf{z}^{(\ell)}, \mathbf{0}).
\end{align}

Here, \(\mathbf{W}^{(\ell)}\) is the weight matrix, \(\mathbf{b}^{(\ell)}\) the bias vector, \(\mathbf{z}^{(\ell)}\) the pre-activation vector, and \(\mathbf{a}^{(\ell)}\) the activation vector at layer \(\ell\).
The output layer (\(\ell = L\)) generates the final result through a linear transformation:
\begin{equation}
  f(\mathbf{x}) = \mathbf{z}^{(L)} = \mathbf{W}^{(L)}\mathbf{a}^{(L-1)} + \mathbf{b}^{(L)}, \quad \mathbf{W}^{(L)} \in \mathbb{R}^{m \times n_{L-1}}, \ \mathbf{b}^{(L)} \in \mathbb{R}^m.
\end{equation}
Classification is performed by selecting the maximum component of the output:
\begin{equation}
  k^* = \arg\max_{1 \leq k \leq m} \left[f(\mathbf{x})\right]_k.
\end{equation}

\subsection{Robustness Verification via Mathematical Optimisation.}\label{r22}


To verify the classification robustness of feedforward ReLU networks under input perturbations, we define the following mathematical optimization problem:
\begin{equation}\label{ct5}
\gamma^* \;=\; \min_{\mathbf{a}^{(0)},\, \mathbf{z}^{(1)}, \ldots, \mathbf{z}^{(L)},\, \mathbf{a}^{(1)}, \ldots, \mathbf{a}^{(L)}} \; \mathbf{c}^{\top}\mathbf{z}^{(L)},
\end{equation}
where \(\mathbf{c}^{\top}=\mathbf{e}^{(k^*)}-\mathbf{e}^{(k)}\) encodes the margin between the correct class \(k^*\) and a competitor \(k\) \cite{batten2021efficient}. Equivalently, one may append an auxiliary affine  layer \(\mathbf{y}=W\,\mathbf{z}^{(L)}\) with rows \(W_j^{\top}=\mathbf{e}^{(k^*)}-\mathbf{e}^{(j)}\) \((j\neq k^*)\), so that \(\min \mathbf{c}^{\top}\mathbf{z}^{(L)}=\min_j y_j\). For notational economy, we therefore drop \(\mathbf{c}^{\top}\) and write the unified objective as
\begin{equation}\label{equa1}
\gamma^{*} \;=\; \min_{\mathbf{a}^{(0)},\, \mathbf{z}^{(1)}, \mathbf{a}^{(1)}, \ldots, \mathbf{z}^{(\ell)}, \mathbf{a}^{(\ell)}} \; \mathbf{z}^{(\ell)}, \quad \ell\in\{1,\ldots,L\}\ \text{or the auxiliary layer},
\end{equation}
which is mathematically equivalent to \eqref{ct5}.
The constraints are imposed as follows:
\begin{enumerate}
    \item \textbf{Input perturbation:}
    \begin{equation}
    \left\| \mathbf{a}^{(0)} - \bar{\mathbf{x}} \right\|_\infty \leq \epsilon,
    \end{equation}
    ensuring the perturbed input lies within the \( l_\infty \)-neighborhood of the original input \( \bar{\mathbf{x}} \).

    \item \textbf{Hidden layer propagation (\( \ell = 1, \ldots, L-1 \)):}
    \begin{equation}
    \mathbf{z}^{(\ell)} = \mathbf{W}^{(\ell)} \mathbf{a}^{(\ell-1)} + \mathbf{b}^{(\ell)}, \quad \mathbf{a}^{(\ell)} = \max \left( 0, \mathbf{z}^{(\ell)} \right).
    \end{equation}

    \item \textbf{Output layer propagation (\( \ell = L \)):}
    \begin{equation}
    \mathbf{z}^{(L)} = \mathbf{W}^{(L)} \mathbf{a}^{(L-1)} + \mathbf{b}^{(L)}.
    \end{equation}
\end{enumerate}

\noindent\textbf{Robustness Conclusion:} If the optimal value \( \gamma^* > 0 \), then for any perturbed input satisfying \( \left\| \mathbf{a}^{(0)} - \bar{\mathbf{x}} \right\|_\infty \leq \epsilon \), the output always satisfies \( z_{k^*}^{(L)} > z_k^{(L)} \) for all \( k \neq k^* \). This proves that the network's prediction for class \( k^* \) is robust under the given input perturbation.


\subsection{ReLU Activation State Classification and Binary Variable Transformation}\label{sec:relu}

The activation state of a ReLU neuron can be classified into three categories based on its input interval \( \mathcal{I}^{(i)} \):
\begin{enumerate}
    \item \textbf{Stable Activation} (\( \mathcal{I}^{(i)} \subseteq \mathbb{R}^+ \)): The output equals the input, requiring no binary variables.
    \item \textbf{Stable Suppression} (\( \mathcal{I}^{(i)} \subseteq \mathbb{R}^- \)): The output is zero, directly expressible via linear constraints.
    \item \textbf{Undetermined State} (\( \mathcal{I}^{(i)} \cap 0 \neq \emptyset \)): Requires mixed-integer encoding to handle non-convexity, quantified by \( \rho^{(i)} \in [0, 1] \).
\end{enumerate}

For unstable ReLU neurons, the piecewise-linear behavior can be modeled in a Mixed-Integer Linear Programming (MILP) framework by introducing binary variables \( q_i^{(\ell)} \in \{0,1\} \) and the following constraints. Let \( z_i^{(\ell)} \) denote the input to neuron \( i \) in layer \( \ell \), and \( a_i^{(\ell)} = \text{ReLU}(z_i^{(\ell)}) \) represent its activation value. The mathematical formulation \cite{tjeng2017evaluating} is:
\begin{equation}
\begin{aligned}
& a_i^{(\ell)} \geq z_i^{(\ell)}, && a_i^{(\ell)} \geq 0, &&& q_i^{(\ell)} \in \{0,1\}\\
& a_i^{(\ell)} \leq z_i^{(\ell)} + M(1 - q_i^{(\ell)}), && a_i^{(\ell)} \leq M  q_i^{(\ell)} 
\end{aligned}
\label{eq:relu_constraints}
\end{equation}

\( M =  \max(\lvert l_i^{(\ell)}\rvert, \lvert u_i^{(\ell)}\rvert) \), where \( l_i^{(\ell)} \) and \( u_i^{(\ell)} \) are precomputed lower and upper bounds of \( z_i^{(\ell)} \).

\section{Formal Foundations of SART}
In this section, we first define a symbolic system to support weight sign decomposition and absolute value encoding. Based on this, we perform an equivalent reformulation of the original verification problem, which serves as a key intermediate step toward binary variable reduction.

\subsection{Symbolic System for Sign-Aware Decomposition}\label{sec:definitions}

\textbf{Weights Sign Decomposition and Mask Rules.} 
The non - negative weight matrix $\mathbf{W}^{(\ell)}_{+} \in \mathbb{R}^{n_\ell\times n_{\ell - 1}}$ and non - positive weight matrix $\mathbf{W}^{(\ell)}_{-}$ are defined as:
\begin{equation}
    \left[\mathbf{W}^{(\ell)}_{+}\right]_{i,j}=\max\left(\mathbf{W}^{(\ell)}_{i,j},0\right),\quad\left[\mathbf{W}^{(\ell)}_{-}\right]_{i,j}=\min\left(\mathbf{W}^{(\ell)}_{i,j},0\right).
\end{equation}
Their binary masks satisfy:
\begin{equation}
\left[\text{Mask}(\mathbf{W}_+^{(\ell)})\right]_{i,j} = \begin{cases} 
1, & \mathbf{W}_{+,i,j}^{(\ell)} > 0 \\
0, & \text{otherwise}
\end{cases}, \quad
\left[\text{Mask}(\mathbf{W}_-^{(\ell)})\right]_{i,j} = \begin{cases} 
1, & \mathbf{W}_{-,i,j}^{(\ell)} < 0 \\
0, & \text{otherwise}
\end{cases}.
\end{equation}
The bias vector are given by:
\begin{equation}
    \mathbf{b}^{(\ell)}_{+}=\max\left(\mathbf{b}^{(\ell)},0\right),\quad\mathbf{b}^{(\ell)}_{-}=\min\left(\mathbf{b}^{(\ell)},0\right).
\end{equation}

\noindent\textbf{Activation and Pre-activation Decomposition.} 
We define the symbolic representations of the pre-activation components \(\mathbf{z}_+^{(\ell-1)}\), \(\mathbf{z}_-^{(\ell-1)}\) and the activation components \(\mathbf{a}_+^{(\ell-1)}\), \(\mathbf{a}_-^{(\ell-1)}\). 
These symbols are not directly computed; instead, they are defined through a mask-based selection applied to the current layer elements, guided by the positive and negative polarity of the weights $\mathbf{W}^{(\ell)}$ from the subsequent layer.
Given the complete activation vector $\mathbf{a}^{(\ell-1)}$ from layer $\ell-1$, its positive and negative components are defined via a masking operation as follows:
\begin{equation}
\begin{split}
\mathbf{a}_+^{(\ell-1, j)} = \left( \mathbf{a}^{(\ell-1)} \cdot \mathbf{1}^\top \odot \left[ \operatorname{Mask}(\mathbf{W}_+^{(\ell)}) \right]^\top \right)[:, j],  \\
\mathbf{a}_-^{(\ell-1, j)} = \left( \mathbf{a}^{(\ell-1)} \cdot \mathbf{1}^\top \odot \left[ \operatorname{Mask}(\mathbf{W}_-^{(\ell)}) \right]^\top \right)[:, j].
\end{split}
\label{eq:combined}
\end{equation}



This operation broadcasts~\cite{harris2020array} $\mathbf{a}^{(\ell-1)}$ to match the dimensions of the weight mask matrix $\operatorname{Mask}(\mathbf{W}^{(\ell)})$, and performs element-wise multiplication with the transposed mask. The resulting matrix's $j$-th column vector corresponds to the input activation vector associated with the $j$-th output neuron.

\begin{figure}[htbp]
    \centering
    \includegraphics[width=0.8\linewidth]{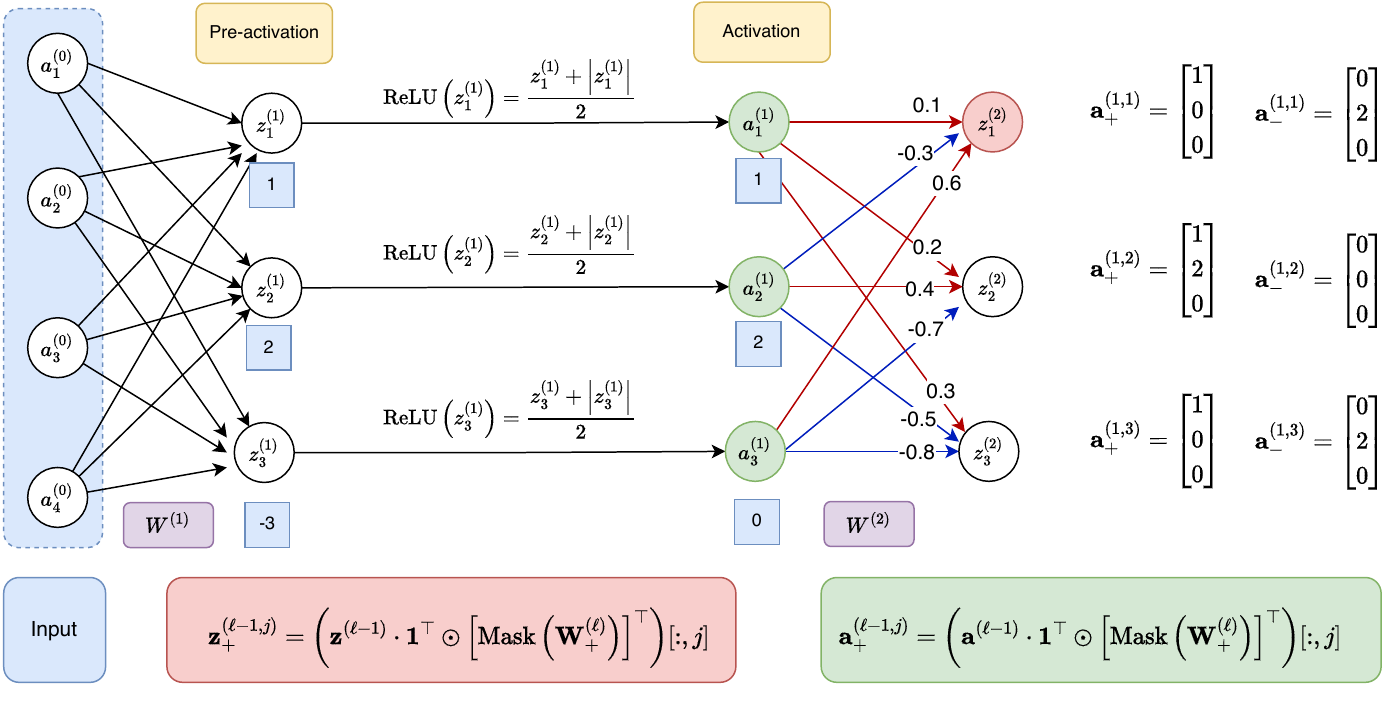}
    \caption{An example defining pre-activation and activation components via weight masking from subsequent layers, used solely for symbolic representation to capture the effect of weight sign.
    (see the supplementary material for details)}
    \label{fig1}
\end{figure}

As shown in Figure~\ref{fig1}, $\mathbf{a}_+^{(1, 1)}$ is defined by the positive weight connections of the first neuron in $\mathbf{W}^{(2)}$, and the result depends entirely on the original values and the sign characteristics of the weights.
Similarly, given the complete pre-activation vector $\mathbf{z}^{(\ell-1)}$ from layer $\ell-1$, its positive and negative components are defined via the masking operation as follows:
\begin{equation}
\begin{split}
\mathbf{z}_+^{(\ell-1, j)} = \left( \mathbf{z}^{(\ell-1)} \cdot \mathbf{1}^\top \odot \left[ \operatorname{Mask}(\mathbf{W}_+^{(\ell)}) \right]^\top \right)[:, j],  \\
\mathbf{z}_-^{(\ell-1, j)} = \left( \mathbf{z}^{(\ell-1)} \cdot \mathbf{1}^\top \odot \left[ \operatorname{Mask}(\mathbf{W}_-^{(\ell)}) \right]^\top \right)[:, j].
\end{split}
\end{equation}

It should be emphasized that when element values are unknown, mask-based selection only defines symbolic structure and cannot infer concrete values, as it operates directly on the elements themselves rather than establishing computational relationships.
This symbolic system partitions activations and pre-activations based on weight signs, serving as the static foundation for the layer-wise recursion in Section~\ref{sec:reform}, enabling absolute value encoding to construct a complete symbolic constraint system starting from the input layer.

\noindent\textbf{ReLU Reformulation via Absolute Value Encoding.} 
The ReLU activation function is decomposed as:
\begin{equation}
    \mathbf{a}^{(\ell)}=\text{ReLU}\left(\mathbf{z}^{(\ell)}\right)=\frac{\mathbf{z}^{(\ell)}+\vert\mathbf{z}^{(\ell)}\vert}{2}.
\end{equation}
Because when \(z^{(\ell)}\) is positive, \(\mathrm{ReLU}\!\left(z^{(\ell)}\right)=\frac{z^{(\ell)}+z^{(\ell)}}{2}=z^{(\ell)}\); when \(z^{(\ell)}\) is negative, \(\mathrm{ReLU}\!\left(z^{(\ell)}\right)=\frac{z^{(\ell)}-z^{(\ell)}}{2}=0\).
The pre-activation term $\mathbf{z}^{(\ell)}$ and its absolute value are computed via:
\begin{align}
    \mathbf{z}^{(\ell)}&=\mathbf{z}^{(\ell)}_{+}+\mathbf{z}^{(\ell)}_{-}=\mathbf{W}^{(\ell)}_{+}\mathbf{a}^{(\ell - 1)}_{+}+\mathbf{W}^{(\ell)}_{-}\mathbf{a}^{(\ell - 1)}_{-}+\mathbf{b}^{(\ell)},\\
    \vert\mathbf{z}^{(\ell)}\vert&=\vert\mathbf{z}^{(\ell)}_{+}+\mathbf{z}^{(\ell)}_{-}\vert=\vert\mathbf{z}^{(\ell)}_{+}\vert+\vert\mathbf{z}^{(\ell)}_{-}\vert \quad\text{(By definition)}.
\end{align}
Extending this formulation to individual neurons, for the \(j\)-th output neuron in layer \(\ell\), the positive-weighted component and its corresponding bias term are given by:  
\begin{equation}
z_{+j}^{(\ell)} = \mathbf{W}_{+j,:}^{(\ell)} \cdot \mathbf{a}_{+}^{(\ell-1,j)} + b_{+j}^{(\ell)}
\end{equation} 
where \(\mathbf{W}_{+j,:}^{(\ell)} \in \mathbb{R}^{n_{\ell-1}}\) denotes the \(j\)-th row of \(\mathbf{W}_{+}^{(\ell)}\), \(b_{+j}^{(\ell)}\) is the corresponding bias component, and \(\mathbf{a}_{+}^{(\ell-1,j)} \in \mathbb{R}^{n_{\ell-1}}\) represents the sparse activation vector aligned with these positive weights. The negative weight components follow an analogous formulation.

\subsection{Intermediate Reformulation via Sign-Absolute Decomposition}\label{sec:reform}
Based on the symbolic system in Section~\ref{sec:definitions}, we construct an intermediate reformulation under the SART framework, which retains all absolute value terms and binary variables as a structural basis for subsequent elimination.

\noindent\textbf{Reformulated Objective Function.}
The objective function in the sign-absolute reformulation of problem~\eqref{equa1} is given by:
\begin{equation}
\gamma_{\min}^{*} = \min_{\substack{\mathbf{a}_+^{(0)},\mathbf{a}_-^{(0)}, \mathbf{z}_+^{(1)}, \mathbf{z}_-^{(1)},\\ \ldots,\mathbf{z}_+^{(L - 1)}, \mathbf{z}_-^{(L - 1)}}} 
\begin{bmatrix}
\frac{1}{2} \left( \mathbf{W}_+^{(\ell)} \mathbf{z}_+^{(\ell - 1)} + \mathbf{W}_-^{(\ell)} \mathbf{z}_-^{(\ell - 1)} \right)\\
+ \frac{1}{2} \left( \mathbf{W}_+^{(\ell)} \left|\mathbf{z}_+^{(\ell - 1)}\right| + \mathbf{W}_-^{(\ell)} \left|\mathbf{z}_-^{(\ell - 1)}\right| \right)\\
+ \mathbf{b}^{(\ell)}
\end{bmatrix}
\label{equa2}
\end{equation}

\noindent\textbf{Reformulated Forward-Propagation Constraints.}
For each sign branch \(s\in\{+,-\}\), the forward propagation on layers \(k=1,\dots,L\) is
\[
\mathbf{z}_{s}^{(k)}=\mathbf{W}_{s}^{(k)}\,\mathbf{a}_{s}^{(k-1)}+\mathbf{b}_{s}^{(k)}, 
\qquad 
\mathbf{a}_{s}^{(k)}=\operatorname{ReLU}\!\big(\mathbf{z}_{s}^{(k)}\big)
=\frac{\mathbf{z}_{s}^{(k)}+\big|\mathbf{z}_{s}^{(k)}\big|}{2},
\]
with known input activation \(\mathbf{a}_{s}^{(0)}\) determined by the input domain.
Eliminating the intermediate activations \(\mathbf{a}_{s}^{(k)}\) by substitution yields a \emph{layer-wise absolute-value encoding}:
\begin{equation}\label{eq26}
\begin{cases}
\mathbf{z}_{s}^{(1)}=\mathbf{W}_{s}^{(1)}\,\mathbf{a}_{s}^{(0)}+\mathbf{b}_{s}^{(1)},\\[4pt]
\mathbf{z}_{s}^{(k)}=\mathbf{W}_{s}^{(k)}\;\dfrac{\mathbf{z}_{s}^{(k-1)}+\big|\mathbf{z}_{s}^{(k-1)}\big|}{2}+\mathbf{b}_{s}^{(k)}, 
\quad k=2,\dots,L.
\end{cases}
\end{equation}
After obtaining \(\mathbf{z}_{s}^{(k)}\), the corresponding activations follow from
\(\mathbf{a}_{s}^{(k)}=\dfrac{\mathbf{z}_{s}^{(k)}+\big|\mathbf{z}_{s}^{(k)}\big|}{2}\).

This reformulation has two properties:
\begin{enumerate}
\item Each layer constraint is standalone (no nested absolute values), so the \(L\) constraints are mutually independent;
\item Every constraint only couples adjacent layers \((k-1)\) and \(k\), preserving the network's layerwise structure.
\end{enumerate}
The same expressions apply simultaneously to both branches by the index \(s\in\{+,-\}\).

\noindent\textbf{Maximization Reformulation.}
To compute neuron upper bounds, we solve the maximization form of the original objective:
\begin{equation}
\max \gamma^{*}=\min(-\mathbf{z}^{(\ell)}).
\end{equation}
After applying weight sign decomposition, the maximization form is reformulated as:
\begin{equation}
\gamma_{\max}^{*}=\min_{\substack{\mathbf{a}_+^{(0)},\mathbf{a}_-^{(0)},\mathbf{z}_{+}^{(1)},\mathbf{z}_{-}^{(1)} \\ \cdots,\mathbf{z}_{+}^{(L - 1)}, \mathbf{z}_{-}^{(L - 1)}}}
\begin{bmatrix}
-\frac{1}{2}\mathbf{W}_{+}^{(\ell)}\mathbf{z}_{+}^{(\ell - 1)}-\frac{1}{2}\mathbf{W}_{-}^{(\ell)}\mathbf{z}_{-}^{(\ell - 1)}\\
-\frac{1}{2}\mathbf{W}_{+}^{(\ell)}\left|\mathbf{z}_{+}^{(\ell - 1)}\right|-\frac{1}{2}\mathbf{W}_{-}^{(\ell)}\left|\mathbf{z}_{-}^{(\ell - 1)}\right|\\
-\mathbf{b}^{(\ell)}
\end{bmatrix}.
\label{equa3}
\end{equation}
The associated constraints remain identical to those of the minimization case.

\section{Sign and Absolute Term Analysis for Binary Variable Reduction}
In this section, we first analyze how coefficient signs affect linearization when only the objective contains absolute values. We then extend the analysis to cases where both the objective and constraints involve absolute values. Finally, leveraging this insight, we reduce binary variables by eliminating certain absolute value terms in the intermediate reformulation problem.

\subsection{Absolute Value Terms in the Objective Function Only}

\subsubsection{Positive Coefficients}\label{411}
Consider the optimization problem \( P_{a} \) proposed in \cite{xu1987special}:
\begin{equation}\label{eq:P_alpha}
\begin{aligned}
(P_a) \quad \min_{X} \quad & f(X) = \textstyle\sum_{i=1}^{n} c_i x_i + \textstyle\sum_{k=1}^{K} \left| \sum_{i=1}^{n} d_{ki} x_i \right| \\
\text{s.t.} \quad & AX = b, \quad X \geq 0.
\end{aligned}
\end{equation}
where \( X = (x_1, x_2, \ldots, x_n) \). To eliminate nonlinearity, \cite{xu1987special} introduces auxiliary variables:
\begin{equation}\label{eq:uv_def}
u_k = \frac{ \left| \sum_{i=1}^{n} d_{ki} x_i \right| - \sum_{i=1}^{n} d_{ki} x_i }{2}, \quad 
v_k = \frac{ \left| \sum_{i=1}^{n} d_{ki} x_i \right| + \sum_{i=1}^{n} d_{ki} x_i }{2}
\end{equation}
yielding an equivalent linear program \( Q_{a} \) and $T_a$ (Complementary Relaxation):

\begin{equation}\label{T_a}
\begin{aligned}
 (T_a) \quad \min_{X,U,V} \quad &  \textstyle\sum_{i=1}^{n} c_{i}x_{i} + \sum_{k=1}^{K} (u_{k} + v_{k}) \\
 \text{s.t.} \quad & AX = b,\quad X \geq 0,\quad U \geq 0,\quad V \geq 0, \\
 & \textstyle\sum_{i=1}^{n} d_{ki}x_{i} + u_{k} - v_{k} = 0, \quad \forall k \in \{1, \ldots, K\}, \\
 & u_{k} \cdot v_{k} = 0, \quad \forall k \in \{1, \ldots, K\} \quad \text{(complementarity condition)}. \\
\end{aligned}
\end{equation}

\begin{equation}\label{eq:Q_alpha}
\begin{aligned}
(Q_{a}) \quad \min_{X,U,V} \quad & \textstyle\sum_{i=1}^{n} c_i x_i + \sum_{k=1}^{K} (u_k + v_k) \\
\text{s.t.} \quad & AX = b, \quad X \geq 0,\quad U \geq 0,\quad V \geq 0,  \\
& \textstyle\sum_{i=1}^{n} d_{ki} x_i + u_k - v_k = 0, \quad \forall k \in \{1, \ldots, K\}. \\
\end{aligned}
\end{equation}

\noindent The key conclusions are established by the following theorems (\Cref{qt,tp}); 
detailed proofs are provided in the supplementary material.


\begin{theorem}\label{qt} 
If \((X^*, U^*, V^*)\) is the optimal solution to problem $Q_a$, then it satisfies the complementary slackness condition \(u_k^* \cdot v_k^* = 0\) for all \(k = 1, 2, \ldots, K\), and it is also the optimal solution to problem $T_a$.
\end{theorem}

\begin{theorem}\label{tp} 
If \( (X', U', V') \) is an optimal solution to Problem $T_a$, then \( X' \) is an optimal solution to Problem $P_a$.
\end{theorem}

\begin{corollary}\label{corollary33}
The optimal solution of the original problem \(P_a\) can be obtained by directly solving the relaxation problem \(Q_a\).
\end{corollary}

\noindent \textbf{Remarks.} By introducing the orthogonality constraint $UV^T = 0$ to define $T_a$, \cite{xu1987special} demonstrates that the reformulated model possesses two important properties. First, the optimal solution of $Q_a$ can be directly translated into an optimal solution of $T_a$, thereby providing an equivalent solution to the original problem $P_a$. Second, the reformulation significantly reduces the computational complexity from exponential $O(2^K)$ to linear $O(K)$. Notably, when the coefficients of the absolute value terms are positive, the optimal solution of $Q_a$ naturally satisfies the complementarity conditions, allowing an equivalent linear programming formulation of $P_a$ to be constructed without introducing any binary variables.

\subsubsection{Negative Coefficients}\label{412}
For the modified problem \( P_s \) with negative coefficients:
\begin{equation}\label{eq:P_s}
\begin{aligned}
(P_s) \quad \min_{X} \quad & f(X) = \textstyle\sum_{i=1}^{n} c_i x_i - \textstyle\sum_{k=1}^{K} \left| \textstyle\sum_{i=1}^{n} d_{ki} x_i \right| \\
\text{s.t.} \quad & AX = b, \quad X \geq 0. 
\end{aligned}
\end{equation}

similar variable substitutions generate \( T_s \) and \( Q_s \):

\begin{equation}\label{eq:T_s}
\begin{aligned}
(T_s) \quad \min_{X,U,V} \quad & \textstyle\sum_{i=1}^{n} c_i x_i - \textstyle\sum_{k=1}^{K} (u_k + v_k) \\
\text{s.t.} \quad & \text{Constraints identical to } T_{a} \\ & (\text{including } UV^T = 0)
\end{aligned}
\end{equation}

\begin{equation}\label{eq:Q_s}
\begin{aligned}
(Q_s) \quad \min_{X,U,V} \quad & \textstyle\sum_{i=1}^{n} c_i x_i - \textstyle\sum_{k=1}^{K} (u_k + v_k) \\
\text{s.t.} \quad & \text{Constraints identical to } Q_{a}
\end{aligned}
\end{equation}

\noindent 
The key conclusions are established by the following theorems (\Cref{qst,tsp}); 
detailed proofs are provided in the supplementary material.

\begin{theorem}\label{qst}
There exists a case where the optimal solution of problem $Q_s$ does not lie within the feasible region of problem $T_s$.
\end{theorem}

\begin{theorem}\label{tsp}
If \((X^*, U^*, V^*)\) is the optimal solution to the complementary relaxation problem $T_s$, then \(X^*\) must be the optimal solution to the original problem $P_s$.
\end{theorem}

%

\begin{corollary}\label{corollary36}
The relaxation \( Q_s\) cannot guarantee the optimal solution to \( P_s\). It is necessary to introduce binary variables to explicitly encode the complementarity conditions.
\end{corollary}

\noindent \textbf{Remarks.} When the coefficients of the absolute value terms are negative, the problem's nature fundamentally changes. First, the solution space becomes separated, causing the optimal solutions of $Q_s$ to no longer necessarily satisfy the complementary slackness condition $UV^T = 0$, and thus $Q_s$'s solution space is no longer contained within that of $T_s$. Second, although the optimal solutions of $T_s$ still correspond to those of $P_s$, the solution sets of $Q_s$ and $P_s$ are no longer equivalent. This discrepancy mainly arises because the negative coefficients reverse the optimization direction, making the objective function tend to maximize the absolute value terms, which conflicts with the minimization objective under positive coefficients. As a result, linear relaxation fails to cover all feasible solutions. To accurately capture the nonconvex constraints and restore equivalence, it is necessary to explicitly enforce the complementary slackness conditions by introducing a binary variable for each absolute value term. This transforms the problem into a mixed-integer linear program (MILP), maintaining an exponential complexity of order $O(2^K)$. Therefore, binary variables are indispensable for handling absolute value terms with negative coefficients.

\subsection{Objective and Constraint Functions Containing Absolute Value Terms}
We consider the following optimization problem, where both the objective function and the constraints contain absolute value terms.
\begin{equation}
\begin{aligned}
(P_a^{full}) \quad \min_{X} \quad & f(X) = \textstyle\sum_{i = 1}^{n} c_{i}x_{i} + \textstyle\sum_{k = 1}^{K} \left| \textstyle\sum_{i = 1}^{n} d_{ki}x_{i} \right| \\
\text{s.t.} \quad & AX = b, \quad X \geq 0, \\
& \textstyle\sum_{i = 1}^{n} g_{i}^{(t)}x_{i} + \textstyle\sum_{l = 1}^{L_{t}} \left| \sum_{i = 1}^{n} h_{li}^{(t)}x_{i} \right| = q^{(t)}, \quad \forall t \in \{1, \ldots, T\} \\
\end{aligned}
\end{equation}

\begin{theorem}\label{yuesu}
Consider the equality constraint of the form
$
\sum_{i=1}^n g_i^{(t)} x_i + \sum_{l=1}^{L_t} \left| \sum_{i=1}^n h_{li}^{(t)} x_i \right| = q^{(t)},  h_{li}^{(t)} \neq 0,  q^{(t)} \in \mathbb{R}.
$
The feasible region defined by this constraint is nonconvex, and modeling each absolute value term
$\left| \sum_{i=1}^n h_{li}^{(t)} x_i \right|$
necessarily requires introducing an independent binary variable \( y_l^{(t)} \in \{0,1\} \). \quad 
\emph{(Detailed proofs are provided in the supplementary material).}
\end{theorem}

\subsubsection{Positive Coefficients}

According to the above theorem, we can derive an equivalent reformulation, denoted as problem \( T_a^{full} \), which is the complete form with complementarity conditions:
\begin{equation}
\begin{aligned}
(T_a^{full}) \quad \min_{X,U,V,W,Y} \quad & \textstyle\sum_{i = 1}^{n} c_{i}x_{i} + \sum_{k = 1}^{K}(u_{k} + v_{k}) \\
\text{s.t.} \quad & AX = b, \quad X \geq 0, \quad 
 \textstyle\sum_{i=1}^{n} d_{ki}x_{i} + u_{k} - v_{k} = 0 , \quad u_{k} \geq 0, \quad v_{k} \geq 0, \quad \forall k \\
& \textstyle\sum_{i = 1}^{n} g_{i}^{(t)}x_{i} + \textstyle\sum_{l = 1}^{L_{t}} w_{l}^{(t)} = q^{(t)}, \quad y_{l}^{(t)} \in \{0, 1\}, \quad w_{l}^{(t)} \geq 0, \quad \forall l, t \\
& w_{l}^{(t)} \geq \textstyle\sum_{i = 1}^{n} h_{li}^{(t)}x_{i}, \quad w_{l}^{(t)} \geq -\textstyle\sum_{i = 1}^{n} h_{li}^{(t)}x_{i}, \quad \forall l, t \\
& \textstyle\sum_{i = 1}^{n} h_{li}^{(t)}x_{i} \leq M_{l}^{(t)} y_{l}^{(t)}, \quad -\textstyle\sum_{i = 1}^{n} h_{li}^{(t)}x_{i} \leq M_{l}^{(t)} (1 - y_{l}^{(t)}), \quad \forall l, t \\
& u_{k} \cdot v_{k} = 0, \quad \forall k \in \{1, \ldots, K\} \quad \text{(complementarity condition)}.
\end{aligned}
\end{equation}
The variable $w_l^{(t)}$ is the sign-replacement representation of the absolute-value term  $ \left|\sum_{i=1}^{n}h_{li}^{(t)}x_{i}\right| $. $M_l^{(t)}$ denotes the big-M constant, chosen no smaller than the absolute values of the upper and lower bounds of $w_l^{(t)}$. 
Removing the complementarity conditions \( u_k \cdot v_k = 0 \) from problem \( T_a^{full} \), we obtain the relaxed formulation \( Q_a^{full} \), where all other constraints remain unchanged.
\begin{equation}
\begin{aligned}
(Q_a^{full}) \quad \min_{X,U,V,W,Y} \quad & \textstyle\sum_{i=1}^n c_i x_i + \sum_{k=1}^K (u_k + v_k) \\
\text{s.t.} \quad 
& \text{same constraints as in } T_a^{\text{full}}, \\
& \text{ except for the removal of } u_k \cdot v_k = 0.
\end{aligned}
\end{equation}
For the problems where both the objective function and the constraints contain absolute value terms, we derive the following theorems based on 
~\Cref{qt,tp,yuesu}:

\begin{theorem}
Let \((X^*, U^*, V^*, W^*, Y^*)\) be an optimal solution of \(Q_a^{full}\). Then it satisfies the complementarity condition \(u_k^* \cdot v_k^* = 0\) for all \(k = 1, \dots, K\), and simultaneously forms an optimal solution to the complementarity problem \(T_a^{full}\).
\end{theorem}

\begin{theorem}
If \((X', U', V', W', Y')\) is an optimal solution to the complementarity problem \(T_a^{full}\), then \(X'\) is an optimal solution to the original problem \(P_a^{full}\).
\end{theorem}

\begin{corollary}\label{corollary310}
The optimal solution of the original problem \(P_a^{full}\) can be obtained by directly solving the relaxation problem \(Q_a^{full}\).
\end{corollary}

\noindent \textbf{Remarks.}
The above theorems show that the optimal solution of the relaxation problem \(Q_a^{full}\) automatically satisfies the complementarity condition associated with the absolute value terms in the objective function, and directly corresponds to the optimal solution of the original problem. The absolute value terms in the objective are precisely modeled using linear complementarity variables \((u_k, v_k)\), without the need for binary variables, while the absolute value terms in the constraints require binary variables \(y_l^{(t)}\) to ensure completeness.

\subsubsection{Negative Coefficients}
Similarly, for the case where both the objective function and the constraint functions contain absolute value terms, but the coefficients of the absolute value terms in the objective function are negative, the problem is formalized as follows:
\begin{equation}
\begin{aligned}
(P_s^{full}) \quad \min_{X} \quad &f(X) = \textstyle\sum_{i = 1}^{n} c_{i}x_{i} - \sum_{k = 1}^{K} \left|\sum_{i = 1}^{n} d_{ki}x_{i} \right| \\
\text{s.t.} \quad &AX = b, \quad X \geq 0,  \\
&\textstyle\sum_{i = 1}^{n} g_{i}^{(t)} x_{i} + \textstyle\sum_{l = 1}^{L_{t}} \left|\sum_{i = 1}^{n} h_{li}^{(t)} x_{i} \right| = q^{(t)}, \quad \forall t \in \{1, \ldots, T\} \\
\end{aligned}
\end{equation}

similar variable substitutions generate \( T_s^{full} \) and \( Q_s^{full} \):

\begin{equation}\label{eq:T_s}
\begin{aligned}
(T_s^{full}) \quad \min_{X,U,V,W,Y} \quad & \textstyle\sum_{i=1}^{n} c_i x_i - \sum_{k=1}^{K} (u_k + v_k) \\
\text{s.t.} \quad & \text{Constraints identical to } T_a^{full}
\end{aligned}
\end{equation}

\begin{equation}\label{eq:Q_s}
\begin{aligned}
(Q_s^{full}) \quad \min_{X,U,V,W,Y} \quad & \textstyle\sum_{i=1}^{n} c_i x_i - \sum_{k=1}^{K} (u_k + v_k) \\
\text{s.t.} \quad & \text{Constraints identical to } Q_a^{full}
\end{aligned}
\end{equation}

For the problems in which both the objective function and the constraints contain absolute value terms, the following theorems can be derived based on 
~\Cref{qst,tsp,yuesu}:

\begin{theorem}
When the absolute value term in the objective function has a negative coefficient, there exists an optimal solution to the relaxation problem \( Q_s^{full}\) that does not satisfy the complementarity condition \( u_k \cdot v_k = 0 \), and thus does not lie in the feasible region of the complementarity problem \( T_s^{full}\).
\end{theorem}

\begin{theorem}
If \( (X^*, U^*, V^*, W^*, Y^*) \) is an optimal solution to the complementarity problem \( T_s^{full}\), then \( X^* \) constitutes an optimal solution to the original problem \( P_s^{full}\).
\end{theorem}

\begin{corollary}\label{corollary313}
The relaxation \( Q_s^{full}  \) cannot guarantee the optimal solution to \( P_s^{full}\). It is necessary to introduce binary variables to explicitly encode the complementarity conditions.
\end{corollary}

\noindent \textbf{Remarks.}
In the case of negative coefficients, the optimal solution to the relaxation problem \( Q_s^{full}\) cannot automatically satisfy the complementarity condition, leading to a fundamental discrepancy between its solution space and the feasible region of the complementarity problem. In this case, it is essential to solve the complementarity-constrained problem \( T_s^{full}\) directly to obtain the optimal solution to the original problem. Crucially, the absolute value terms in the objective function require binary variables to enforce the complementarity condition, while the absolute value terms in the constraints still require additional binary variables \( y_l^{(t)} \) for modeling. Therefore, the linearization advantage seen in the case of positive coefficients is no longer attainable, resulting in a fundamental computational bottleneck.

\noindent \textbf{Rules.}
Based on the previous analysis of absolute value terms in the objective function and constraints, this section summarizes the decision rules for introducing binary variables when linearizing non-nested absolute value terms. These rules depend on the optimization direction (minimization or maximization), the position (objective or constraint) of the absolute value terms, and the sign of their coefficients. Specifically, positive coefficient terms in the objective function do not require binary variables in minimization but do in maximization; negative coefficient terms have the opposite requirement. Absolute value terms in constraints require binary variables regardless of coefficient sign and optimization direction. The above conclusions apply to non-nested structures. The rules are summarized in ~\Cref{tab:abs-value-binary}.

\begin{table}[htbp]
\centering
\caption{Binary variable usage for absolute value terms in optimization problems}
\label{tab:abs-value-binary}
\renewcommand{\arraystretch}{1.0}
\begin{tabular}{ccc}
\toprule
\textbf{Type} & \textbf{Minimization} & \textbf{Maximization} \\
\midrule
Positive coeff.\ in objective & No binary vars & Binary vars \\
Negative coeff.\ in objective & Binary vars & No binary vars \\
In constraints (any sign) & Binary vars & Binary vars \\
\bottomrule
\end{tabular}
\end{table}

\subsection{Final Reformulation and Binary Variable Reduction via SART}

In this subsection, based on the above theoretical analysis, we first eliminate certain absolute value terms in the intermediate reformulation and propose an equivalent final reformulation. Then, we analyze the reduction in binary variables associated with unstable neurons in neural network verification, covering both single-layer and multi-layer cases. Finally, we present the SART binary minimality theorem.

\subsubsection{Final Reformulation with Eliminated Certain Absolute Terms}

We perform a variable substitution on the decomposed objective function and constraints in Section~\ref{sec:reform} to eliminate certain absolute value terms.

\paragraph{Lower Bound Computation} The intermediate objective function \eqref{equa2} can be reformulated as:
\begin{equation}
\gamma_{\min}^{*} = \min_{\substack{\mathbf{a}_+^{(0)},\mathbf{a}_-^{(0)},\mathbf{z}_{+}^{(1)}, \mathbf{z}_{-}^{(1)}, \ldots, \\ \mathbf{z}_{+}^{(\ell-1)}, \mathbf{z}_{-}^{(\ell-1)}, \mathbf{u}, \mathbf{v}}} 
\begin{bmatrix} 
\frac{1}{2} \mathbf{W}_+^{(\ell)} (\mathbf{v} - \mathbf{u}) + \frac{1}{2} \mathbf{W}_-^{(\ell)} \mathbf{z}_-^{(\ell-1)} \\
 + \frac{1}{2} \mathbf{W}_+^{(\ell)} (\mathbf{u} + \mathbf{v}) + \frac{1}{2} \mathbf{W}_-^{(\ell)} \left|\mathbf{z}_-^{(\ell-1)}\right| \\ 
 + \mathbf{b}^{(\ell)}
\end{bmatrix}
\end{equation}
By introducing non-negative auxiliary variables $\mathbf{u}$ and $\mathbf{v}$, we add the following constraints:
\begin{equation}
\mathbf{z}_{+}^{(\ell - 1)} + \mathbf{u} - \mathbf{v} = 0, \quad \mathbf{u} \geq 0, \quad \mathbf{v} \geq 0.
\end{equation}
This transformation converts the nonlinear characteristics of activation functions into a combination of linear terms and absolute value terms in the objective function. The separation of weight matrices into positive and negative components ($\mathbf{W}_{+}^{(\ell)}$ and $\mathbf{W}_{-}^{(\ell)}$) simplifies the optimization complexity while maintaining theoretical consistency.

\paragraph{Upper Bound Computation} The intermediate objective function \eqref{equa3} can be restructured as:
\begin{equation}
\gamma_{\max}^{*} = \min_{\substack{\mathbf{a}_+^{(0)},\mathbf{a}_-^{(0)}, \mathbf{z}_{+}^{(1)}, \mathbf{z}_{-}^{(1)},\ldots, \\ \mathbf{z}_{+}^{(\ell - 1)}, \mathbf{z}_{-}^{(\ell - 1)},\mathbf{s}, \mathbf{t}}} \begin{bmatrix} 
-\frac{1}{2} \mathbf{W}_{+}^{(\ell)} \mathbf{z}_{+}^{(\ell - 1)} - \frac{1}{2} \mathbf{W}_{+}^{(\ell)} (\mathbf{t} - \mathbf{s}) \\
-\frac{1}{2} \mathbf{W}_{-}^{(\ell)} \left|\mathbf{z}_{-}^{(\ell - 1)}\right| - \frac{1}{2} \mathbf{W}_{-}^{(\ell)} (\mathbf{s} + \mathbf{t}) \\
-\mathbf{b}^{(\ell)}
\end{bmatrix}
\end{equation}
With non-negative auxiliary variables $\mathbf{s}$ and $\mathbf{t}$, we introduce the constraints:
\begin{equation}
\mathbf{z}_{-}^{(\ell - 1)} + \mathbf{s} - \mathbf{t} = 0, \quad \mathbf{s} \geq 0, \quad \mathbf{t} \geq 0.
\end{equation}
These constraints ensure the feasibility of the decomposition process. The separation of weight matrices into $\mathbf{W}_{+}^{(\ell)}$ and $\mathbf{W}_{-}^{(\ell)}$ enables conversion of specific absolute value terms into linear constraints while preserving theoretical consistency.

\paragraph{Big-M Method for Remaining Absolute Terms} For non-eliminable absolute terms, we employ the big-\(M\) method and introduce binary variables to transform them, where each remaining absolute term corresponds uniquely to a binary variable. 
For each element \( z_i^{(\ell-1)} \), define its absolute value \( y_i^{(\ell-1)} = |z_i^{(\ell-1)}| \), and impose the following constraints:
\begin{equation}
\begin{aligned}
y_{i}^{(\ell-1)} &\geq z_{i}^{(\ell-1)}, && y_{i}^{(\ell-1)} \geq -z_{i}^{(\ell-1)}, && q_{i}^{(\ell-1)} \in \{0,1\},\\
y_{i}^{(\ell-1)} &\leq z_{i}^{(\ell-1)} + 2(1 - q_{i}^{(\ell-1)})M, \quad && y_{i}^{(\ell-1)} \leq -z_{i}^{(\ell-1)} + 2q_{i}^{(\ell-1)}M.  \\
\end{aligned}
\quad \forall i
\end{equation}
\( M =  \max(\lvert l_i^{(\ell-1)}\rvert, \lvert u_i^{(\ell-1)}\rvert) \),  \( l_i^{(\ell-1)} \) and \( u_i^{(\ell-1)} \) are precomputed lower and upper bounds of \( z_i^{(\ell-1)} \).

\subsubsection{Overall Analysis of Reducing Binary Variables}

To rigorously compute the tightest possible bounds ($\gamma_{\min}^{*}$ and $\gamma_{\max}^{*}$) for neural network verification, we systematically analyze the binary variable reduction mechanism. Our analysis assumes a worst-case scenario where the proportion of unstable neurons $\rho^{(m)} = 1$ at each layer $m$. Practical applications should scale the results by the actual $\rho^{(m)}$ values.
In order to distinguish between the number of non - zero weights in the entire weight matrix and the number of non - zero weights associated with unstable neurons, we introduce the following definitions.

\begin{definition}[Nonzero Weight Count]
For a given layer \( \ell \) of a neural network, its weight matrix \( \mathbf{W}^{(\ell)} \) can be decomposed into a non-negative part \( \mathbf{W}_+^{(\ell)} \geq 0 \) and a non-positive part \( \mathbf{W}_-^{(\ell)} \leq 0 \). We define the following terms:
\end{definition}

\begin{itemize}
    \item Global Nonzero Weight Count:  
    Let \( T_+^{(\ell)} \) and \( T_-^{(\ell)} \) denote the total number of nonzero elements in \( \mathbf{W}_+^{(\ell)} \) and \( \mathbf{W}_-^{(\ell)} \), respectively:
    \begin{equation}
        T_+^{(\ell)} = \left| \{ w \mid w \in \mathbf{W}_+^{(\ell)}, w \neq 0 \} \right|, \quad
        T_-^{(\ell)} = \left| \{ w \mid w \in \mathbf{W}_-^{(\ell)}, w \neq 0 \} \right|.
    \end{equation}

    \item Nonzero Weight Count Corresponding to Unstable Neurons: 
%
%
%
For layer $\ell$, let $\rho^{(\ell)}_+$ be the fraction of unstable neurons among those associated with positive weights, and let $\rho^{(\ell)}_-$ be the fraction among those associated with negative weights. Define
\begin{equation}
N^{(\ell)}_+ = \rho^{(\ell)}_+\, T^{(\ell)}_+ , \qquad
N^{(\ell)}_- = \rho^{(\ell)}_-\, T^{(\ell)}_- .
\end{equation}
Boundary cases for the positive side: if $\rho^{(\ell)}_+=1$ then $N^{(\ell)}_+=T^{(\ell)}_+$; if $0<\rho^{(\ell)}_+<1$ then $N^{(\ell)}_+$ counts only the nonzero positive weights of unstable neurons.
For the negative side: if $\rho^{(\ell)}_-=1$ then $N^{(\ell)}_-=T^{(\ell)}_-$; if $0<\rho^{(\ell)}_-<1$ then $N^{(\ell)}_-$ counts only the nonzero negative weights of unstable neurons.
The proportions $\rho^{(\ell)}_+$ and $\rho^{(\ell)}_-$ are computed independently.

\end{itemize}

\subsubsection{Single-Layer Analysis}
For a single-layer neural network, during the transformation of the optimization problem, the key improvements leverage the sign characteristics of the weight matrix:
\begin{enumerate}
    \item Lower bound problem: For non-negative weight processing, when the sign of the coefficient preceding an absolute value term is non-negative, the corresponding absolute value term can be eliminated without introducing additional binary variables. In this case, the number of required binary variables reduces to \(N^{(\ell)}_-\).\label{case1}
    
    \item Upper bound problem: For non-positive weight processing (converted to non-negative variables via sign flipping), the absolute value terms can similarly be removed without extra binary variables, resulting in \(N^{(\ell)}_+\) binary variables.\label{case2}
\end{enumerate}

In the single-layer case, the essence of the problem lies in the fact that only the objective function contains absolute value terms, while the constraints do not. In this case, the problem can be transformed into two types, $P_a$ and $P_s$, based on the sign of the coefficients preceding the absolute value terms in the objective function. 
This transformation is justified by the observation from ~\Cref{eq26} that the output of the final layer $z^{(\ell)}$ actually depends on the previous layer $z^{(\ell-1)}$, allowing the absolute value terms originally appearing in the final layer to be rewritten as absolute value terms in the objective function.

The intuition for eliminating binary variables comes from the convexity of the objective under specific optimization directions and the resulting automatic tightness of the linearization. As proved in \Cref{411} (\Cref{qt,tp}), when minimizing absolute-value terms with nonnegative coefficients, the objective naturally drives the auxiliary variables to their lower bounds, so exact equivalence can be preserved without binary variables. Both the lower-bound problem in case(\ref{case1}) and case(\ref{case2}), after rewriting the upper bound as minimizing a negated objective, satisfy this condition, so the binaries can be safely removed. In contrast, for the negative-coefficient nonconvex cases in \Cref{412}, the linearization is no longer automatically tight, and binary variables must be retained to ensure mathematical equivalence.

\noindent \textbf{Reduction Analysis.} In the original problem, obtaining the lower and upper bounds of an unstable neuron requires introducing one binary variable for each bound, resulting in a total of \( 2(N^{(\ell)}_+ + N^{(\ell)}_-) \). Through the sign-absolute reformulation, the total count is reduced to \( N^{(\ell)}_+ + N^{(\ell)}_- \), effectively halving the number of binary variables. By eliminating redundant binary variables, the complexity of the mixed-integer programming model is significantly reduced while preserving mathematical equivalence.

\subsubsection{Multi-Layer Analysis}
In the multi-layer case, the essence of the problem lies in the presence of absolute value terms in both the objective function and the constraint functions. Based on the sign of the coefficients preceding the absolute value terms in the objective function, the problem can be transformed into $P_a^{full}$ and $P_s^{full}$.
Similar to the single-layer case, the absolute values in the final layer can still be incorporated into the objective function. Since our model does not contain nested absolute value terms, the absolute values in the intermediate layers cannot be further transformed and must remain in the constraints.

Consider a network with $L$ layers, where the intermediate layer $m$ ($1 \leq m \leq L - 1$) reduces binary variable requirements through sign decomposition of weight matrices into non-negative components $\mathbf{W}_{+}^{(m)} > 0$ and non-positive components $\mathbf{W}_{-}^{(m)} < 0$. For a subnetwork containing $m$ layers, the processing strategy exhibits layer-specific characteristics. 

The first $m - 1$ layers retain traditional binary variable processing due to nested absolute value terms in activation variables. Layer $m$ as the terminal layer implements sign decomposition with non-negative weights $\mathbf{W}_{+}^{(m)} \geq 0$ acting directly on $|z_{+}^{(m-1)}|$ and non-positive weights $\mathbf{W}_{-}^{(m)} \leq 0$ operating on $|z_{-}^{(m-1)}|$ through sign inversion, thereby reducing binary variable requirements.

Binary variable calculation follows layer-wise patterns. In the initial $m - 1$ layers, each layer $t$ ($t = 1, \ldots, m - 1$) requires dual binary variables per absolute value component, yielding $2(N_{+}^{(t)} + N_{-}^{(t)})$ variables per layer. The cumulative requirement becomes $\sum_{t=1}^{m-1} 2(N_{+}^{(t)} + N_{-}^{(t)})$. The terminal layer $m$ demonstrates optimization: non-negative weights eliminate absolute values directly while non-positive weights achieve equivalent elimination through sign inversion, requiring only $N_{+}^{(m)} + N_{-}^{(m)}$ binary variables corresponding to non-zero weight elements.

Total binary variables exhibit significant reduction through this optimization. The baseline unoptimized configuration requires:
$
\text{Total}_{\text{orig}} = \sum_{t=1}^{m} 2(N_{+}^{(t)} + N_{-}^{(t)}),
$
whereas the optimized implementation reduces to:
$
\text{Total}_{\text{opt}} = \sum_{t=1}^{m-1} 2(N_{+}^{(t)} + N_{-}^{(t)}) + (N_{+}^{(m)} + N_{-}^{(m)}).
$
The resultant variable reduction calculates as:
$
\Delta = \text{Total}_{\text{orig}} - \text{Total}_{\text{opt}} = N_{+}^{(m)} + N_{-}^{(m)}.
$

\noindent \textbf{Reduction Analysis.} (1) The terminal layer's binary variables decrease by 50\% from $2(N_{+}^{(m)} + N_{-}^{(m)})$ to $N_{+}^{(m)} + N_{-}^{(m)}$; (2) The first $m - 1$ layers maintain original requirements; (3) The total reduction equals the number of non-zero weights in layer $m$, specifically $N_{+}^{(m)} + N_{-}^{(m)}$.

\begin{theorem}[Minimality of Binary Variables in SART]
\label{thm:sart-minimality}
Under the assumptions of completeness and soundness, SART achieves the minimal number of binary variables required for unstable neurons. Any further reduction would either violate the assumption \( P \neq NP \) or compromise completeness of the verification.\quad 
\emph{(Detailed proofs are provided in the supplementary material).}
\end{theorem}

This result relies on Propositions~\ref{corollary33}, \ref{corollary36}, \ref{corollary310}, and \ref{corollary313}, which establish the algebraic foundations of the symbolic decomposition and absolute-value modeling framework in SART. These support the Final Reformulation and enable binary variable reduction in both single-layer and multi-layer networks.
In single-layer networks, the Final Reformulation transforms \(P_a\) and \(P_s\) into \(Q_a\) and \(T_s\), halving the binary variables for unstable neurons via a joint sign-absolute reformulation. In multi-layer networks, reformulating \(P_a^{full}\) and \(P_s^{full}\) into \(Q_a^{full}\) and \(T_s^{full}\) allows compression only at the final layer; intermediate layers require one binary variable per unstable neuron due to inter-layer dependencies. Thus, SART uses half the binary variables for compressible output neurons plus one per unstable neuron in intermediate layers, achieving the theoretical minimum under completeness.



\subsubsection{Sign Symmetry and Verification Friendliness}\label{sec:3.4.1}
A mixed-weight neural network allows weights to contain both positive and negative values, and its verification complexity exhibits an exponential relationship with the number of binary variables. By leveraging the weight sign decomposition theory proposed in the previous section and introducing \textbf{sign symmetry constraints} on unstable neurons at each layer, we achieve the theoretically optimal worst-case verification complexity. This provides a theoretical foundation for designing verification-friendly networks.

\begin{definition}[Sign Symmetry]\label{def:sym}
Consider the weight matrix of the layer $m$ in a mixed-weight neural network, decomposed into non-negative and non-positive parts, $W_{+}^{(m)} \geq 0$ and $W_{-}^{(m)} \leq 0$, with total nonzero weights $T^{(m)}$. Given the proportion of unstable neurons caused by positive weights \(\rho_+^{(m)}\) and the proportion caused by negative weights \(\rho_-^{(m)}\), the nonzero weights associated with unstable neurons are defined as:
\[
N_+^{(m)} = \rho_+^{(m)} T_+^{(m)}, \quad N_-^{(m)} = \rho_-^{(m)} T_-^{(m)}.
\]
The layer satisfies sign symmetry if:
\begin{equation}\label{eq:sym_cond}
\left| N_{+}^{(m)} - N_{-}^{(m)} \right| \leq 1.
\end{equation}

Specific conditions include:
\begin{itemize}
\item \textbf{Even allocation}: For even total non-zero weights \(S = N_{+}^{(m)} + N_{-}^{(m)}\):
\begin{equation}\label{eq:even}
N_{+}^{(m)} = N_{-}^{(m)} = S/2.
\end{equation}
\item \textbf{Odd allocation}: For odd \(S\), allow \(\pm 1\) difference:
\begin{equation}\label{eq:odd}
N_{+}^{(m)} = \left\lfloor S/2 \right\rfloor,\quad N_{-}^{(m)} = \left\lceil S/2 \right\rceil \quad \text{(or vice versa)}.
\end{equation}
\end{itemize}
\end{definition}

The verification time complexity for the terminal layer is modeled as:
\begin{equation}\label{eq:complexity}
O\left(2^{N_+^{(m)}} + 2^{N_-^{(m)}}\right). 
\end{equation}

\begin{theorem}[Minimum Complexity Condition]\label{thm:min}
Given fixed \(S = N_{+}^{(m)} + N_{-}^{(m)}\), the time complexity reaches its minimum if and only if sign symmetry (\ref{eq:sym_cond}) is satisfied.\quad
\emph{(Detailed proofs are provided in the supplementary material).}
\end{theorem}


We would like to clarify the distinction between encoding representation and the solving process. For encoding, Mao et al.~\cite{mao2024expressiveness} show that multi-neuron convex relaxations cannot guarantee completeness in theory and incur irreducible approximation error, so a complete formulation must introduce binary variables to capture the underlying disjunctive structure. For solving, after the model is fully encoded, complete methods such as MNBaB~\cite{ferrari2022complete} can integrate multi-neuron relaxations within branch-and-bound(BaB) to reduce the number of binary decisions that need to be handled in practice. This is a dynamic reduction enabled by the solving strategy and does not change the complete static structure required by the encoding . Any efficiency gain in solving presupposes a complete encoding. Theorem 4.17 is intended to emphasize that, given a complete encoding and the same solving algorithm such as LayerABS, networks with sign symmetry enjoy the best worst-case complexity.

\noindent\textbf{Clarification on verification friendliness.} In this paper, ``verification friendliness'' is a rigorous theoretical statement under SART: while preserving equivalence, SART reduces the number of binary variables in the verification encoding, thereby shrinking the search space at the level of theoretical complexity. This does not imply polynomial-time solvability in general; under the assumption $P \neq NP$, the worst case remains hard. We currently have no empirical evidence that sign symmetric networks are easier to verify; such evaluation is valuable but independent of this theoretical claim.

\section{Bad News vs. Good News: Static and Dynamic Perspectives}
Building on the SART framework, we have proved that, while preserving equivalence, imposing absolute-value-related constraints on the network's final layer reduces the required number of binary variables by 50\%. This provides a solid foundation for optimizing the final-layer structure. However, this compression advantage is confined to the final layer and cannot be directly applied to intermediate layers. The reported 50\% reduction refers to the last hidden layer, since the network's output layer typically does not use ReLU activations.

From a static perspective, this may look like bad news for verification---if the entire verification task is treated as a single large MILP, compressing only the last layer has limited impact. Yet the bad news hides good news. Verification is inherently a sequential process of dynamically accumulating information and can be viewed as a series of highly correlated small MILPs whose efficiency is bottlenecked by the search space. The performance of verification algorithms depends critically on obtaining tight variable bounds early in the search tree.
Motivated by this insight, we introduce a SART-based progressive tightening strategy and implement it systematically as LayerABS. Rather than fully encoding the entire network at once, the strategy treats intermediate layers as temporary final layers and applies SART encoding to them, thereby continually compressing the search space. This yields a twofold benefit:
\begin{itemize}
  \item \textbf{Local:} SART achieves a 50\% reduction in binary variables at the temporary final layer, directly improving the efficiency of the current subproblem.
  \item \textbf{Global:} 
  Each window builds on the tightened bounds computed in preceding layers, leading to a cascading tightening of intermediate-layer output ranges, early fixation of many neuron activations, and reduced downstream binary-variable requirements.
\end{itemize}

\begin{figure}[htbp]
    \centering
    \includegraphics[width=0.7\linewidth]{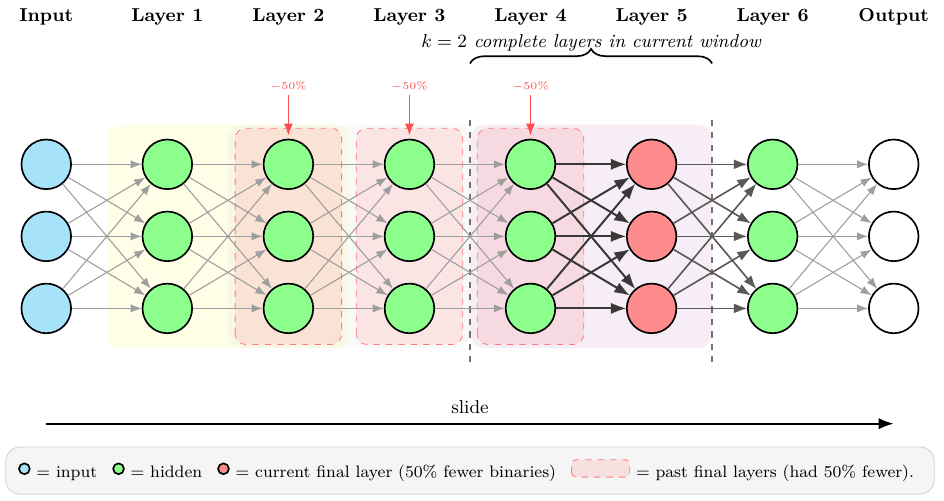}
\caption[Lifting Final-Layer Compression to the Entire Network via Progressive Tightening]{\textbf{Lifting Final-Layer Compression to the Entire Network via Progressive Tightening.}
LayerABS applies SART to a sliding window of \(k\) complete layers, treating intermediate layers as temporary finals; at each step, it halves the binaries at the window tail (50\% fewer at the current final layer) and leverages the tightened bounds from preceding layers forward, shrinking the search space and reducing global binaries.}
    \label{garph_k}
\end{figure}

Through this chain effect in dynamic verification, SART ultimately delivers global acceleration.
At its core, LayerABS integrates symbolic bound propagation, weight-sign decomposition, and absolute-value constraint encoding into a three-level progressive mechanism that systematically compresses both the solution space and the scale of binary variables.

Specifically, LayerABS consists of the following three progressively tightening stages:

\noindent \textbf{Stage 1 --- Abstract Interpretation Encoding.}  
In this stage, all network layers adopt abstract interpretation  methods (e.g., the DeepPoly abstract domain~\cite{singh2019abstract}). By symbolic propagation or linear relaxation techniques, conservative interval bounds \( T^{(i)} \) are generated for each neuron. This stage maps all neurons (including those in uncertain states) into linear inequality constraints, completely avoiding the introduction of binary variables. The purely abstract modeling has low computational overhead and is suitable for quickly filtering out the majority of easily decidable input spaces. Although its precision is limited and may fail to satisfy complex property verification requirements, it provides a good preprocessing foundation for shrinking the verification space in subsequent stages.

\noindent \textbf{Stage 2 --- Hybrid Abstract SART Encoding.}  
If the first stage returns ``unknown'', LayerABS proceeds to the second stage, applying complete mixed-integer constraint encoding to the fixed last \( k \) layers of the network (usually \( k=2 \)), while retaining the abstract constraints from the first stage in the previous \( m-k \) layers. Notably, all complete encodings are based on the SART theory. The difference is that the final layer can exploit the sign-absolute value structure to achieve binary variable compression, while intermediate layers, located at the absolute value terms in the constraint functions, cannot reduce the number of binary variables in theory.
Specifically:  
\begin{itemize}
    \item For the last \( m \)-th layer, weight sign decomposition splits the weight matrix into non-negative part \( W_+^{(m)} \) and non-positive part \( W_-^{(m)} \). Non-negative sign constraints \( z_+^{(m-1)} \geq 0 \), \( z_-^{(m-1)} \geq 0 \) are introduced to linearize the absolute value terms \( | z_+^{(m-1)} | \), compressing the binary variable count in this layer from 
$
2\left(N_+^{(m)} + N_-^{(m)}\right)  \text{to} \left(N_+^{(m)} + N_-^{(m)}\right).
$
    \item Intermediate layers (from layer \( m-k+1 \) to \( m-1 \)) also adopt complete encodings based on SART. However, since they appear as absolute value terms in constraint functions, binary variable compression cannot be achieved; their binary variable count remains
$
  2 \left( N_+^{(\ell)} + N_-^{(\ell)} \right).
$
    \item The previous \( m-k \) layers continue using the abstract linear constraints from stage one without introducing any binary variables, significantly reducing the overall variable scale.
\end{itemize}
Therefore, the total number of binary variables in the second stage is:  
\begin{equation}
B_{\text{opt}} = \sum_{\ell = m-k+1}^{m-1} 2 \left( N_+^{(\ell)} + N_-^{(\ell)} \right) \quad\\+ \left( N_+^{(m)} + N_-^{(m)} \right).
\end{equation}
In contrast, if all layers apply the standard MILP encoding, the number of binary variables is:  
$
B_{\text{orig}} = \sum_{\ell=1}^{m} 2 \left( N_+^{(\ell)} + N_-^{(\ell)} \right).
$
The resulting binary variable reduction is:  
$
\Delta B = \sum_{\ell=1}^{m-k} 2 \left( N_+^{(\ell)} + N_-^{(\ell)} \right) +\sum_{\ell=m-k+1}^m \left( 2 - \delta_{\ell,m} \right) \left( N_+^{(\ell)} + N_-^{(\ell)} \right),
$
where \( \delta_{\ell,m} \) is the Kronecker delta function, taking value 1 when \( \ell = m \) and 0 otherwise.
Since the last layer achieves a theoretical 50\% compression rate, the default choice \( k=2 \) achieves a good balance between variable compression and modeling precision.

This hierarchical refinement strategy not only reduces the overall variable scale but also has inherent structural advantages: the verification process proceeds forward layer by layer, and the tightness of constraints in the earlier layers directly influences the feasible space of later layers, allowing subsequent solving to be performed in a smaller, more focused region and significantly improving overall verification efficiency.

\noindent \textbf{Stage 3 --- Full Complete SART Encoding.}  
If the second stage returns ``unknown'', LayerABS falls back to applying complete SART-based mixed-integer constraint encoding to all layers of the network. Specifically:  
\begin{itemize}
    \item The first \( m-1 \) layers all adopt the full ReLU constraints with SART encoding; their binary variable count cannot be compressed and remains
$
  2 \left( N_+^{(\ell)} + N_-^{(\ell)} \right). 
$
    \item The final \( m \)-th layer continues to apply the sign-absolute value reformulation and linearization, achieving binary variable compression.
\end{itemize}
At this point, the total number of binary variables is:  
\begin{equation}
B_{\text{fallback}} = \sum_{\ell=1}^{m-1} 2 \left( N_+^{(\ell)} + N_-^{(\ell)} \right)  + \left( N_+^{(m)} + N_-^{(m)} \right).
\end{equation}
This fallback mechanism maximizes retention of the structural optimization advantage of the last layer while ensuring verification completeness and avoiding unnecessary performance degradation.

\vspace{0.4cm}
\begin{algorithm}[H]
\caption{LayerABS\\[-0.25em]\footnotesize
Stages 1-2 return $\{\texttt{safe},\texttt{unknown},\texttt{timeout}\}$; Stage 3 returns $\{\texttt{safe},\texttt{unsafe},\texttt{timeout}\}$.}
\label{alg:layerabs_modified}
\DontPrintSemicolon
\SetKwInput{KwIn}{Input}
\SetKwInput{KwOut}{Output}
\SetKwFunction{Verify}{Verify}
\SetKw{Return}{return}

\KwIn{network $f$ with $m$ layers; input $X$; specification $\epsilon$; timeout $T$; terminal layers $k$}
\KwOut{Result $\in \{\texttt{safe},\texttt{unsafe},\texttt{timeout}\}$}

$result \gets$ Verify(AbstractEncoding($f$, $X$, full network), $\epsilon$, $T$); \qquad //stage 1\; 
\If{$result \in \{\texttt{safe},\texttt{timeout}\}$}{
    \Return $result$
}

$result \gets$ Verify(PartialSARTEncoding($f$, $X$, first $m-k$ layers abstract, last $k$ layers complete), $\epsilon$, $T$); \qquad //stage 2\;
\If{$result \in \{\texttt{safe},\texttt{timeout}\}$}{
    \Return $result$
}

$result \gets$ Verify(SARTEncoding($f$, $X$, full network), $\epsilon$, $T$) \qquad //stage 3\;
\Return $result$\;

\end{algorithm}

\section{Evaluation}

\noindent \textbf{Environment Setup.} CPU experiments were conducted on Ubuntu 20.04 with an AMD Ryzen 9 7950X (16 cores/32 threads), 128GB RAM, Python 3.9, and Gurobi 11. GPU experiments used an ESXi-based virtual server with an NVIDIA RTX 3090 (Driver 535.183, CUDA 12.2), an AMD EPYC 9754 (16 cores), and 168GB RAM.


\noindent\textbf{Datasets and Networks.}
We evaluate on MNIST (28\(\times\)28 grayscale) using six fully connected ReLU networks (\(5\times 50\), \(5\times 80\), \(6\times 100\), \(9\times 100\), \(10\times 80\), \(9\times 200\)), each trained and verified on 100 correctly classified images.
We also include the VNN-COMP ERAN benchmark (architectures \(5\times 100\), \(8\times 100\)) and CIFAR-10 (architectures \(5\times 50\), \(6\times 80\)).
All experiments use \(\ell_{\infty}\)-norm perturbations on normalized inputs; each image is paired with a perturbation bound \(\epsilon\) and a target label.


\noindent\textbf{Baselines.}
We compare against state-of-the-art complete verifiers: the MIP-based \textsf{MIPVerify}, the SMT-based \textsf{Marabou}, and the branch-and-bound \textsf{$\beta$-CROWN~BaB} (VNN-COMP winner); and leading incomplete methods: \textsf{PRIMA} (SOTA multi-neuron) and abstract-interpretation approaches \textsf{DeepPoly} and \textsf{$\alpha$-CROWN}.

%

\noindent\textbf{Parameter Configuration.}
Experiments use a 2{,}000s timeout, 28 threads, and \textsf{$\beta$-CROWN BaB} with default settings and \textsf{kFSB} branching strategy~\cite{de2021improved};  complete constraints are applied to the last \(k\) layers with \(k\) chosen conservatively, lower for complete runs and slightly higher for incomplete runs (2 and 3, respectively).


\noindent\textbf{Evaluation Metrics.}
Time is reported in seconds as the average verification duration, and the perturbation magnitude is denoted by \(\epsilon\).
Complete verification returns one of three outcomes: \textit{safe}, \textit{unsafe}, or \textit{timeout}.
Incomplete verification returns \textit{safe}, \textit{unknown}, or \textit{timeout}.
In Table~\ref{tab:verification}, single-value entries in the \textit{safe/unsafe} row indicate the number of \textit{safe} cases.

\subsection{Performance of Pure SART Encoding}\label{pure}

\begin{table}[htbp]
\centering
\caption{Pure SART vs.\ Standard MILP (full complete, no abstraction).
Time (w/ to): mean including timeouts at the time limit; Time (excl.): mean over non-timeouts only.}

\label{no_abstraction1}
\begingroup
\setlength{\tabcolsep}{5pt}        
\renewcommand{\arraystretch}{1.2}    
\resizebox{\textwidth}{!}{%
\begin{tabular}{@{}lcccccccc@{}}
\toprule
\multirow{2}{*}{Model}
& \multicolumn{4}{c}{Standard MILP (No Abstraction)}
& \multicolumn{4}{c}{\textbf{Our SART (No Abstraction)}} \\
\cmidrule(lr){2-5} \cmidrule(lr){6-9}
& Safe | Unsafe $_{sum\uparrow}$ & Timeout $_{\downarrow}$ & Time (w/ to) $_{\downarrow}$ & Time (excl.) $_{\downarrow}$
& \textbf{Safe | Unsafe $_{sum\uparrow}$} & \textbf{Timeout $_{\downarrow}$} & \textbf{Time (w/ to) $_{\downarrow}$} & \textbf{Time (excl.) $_{\downarrow}$} \\
\midrule
\makecell[l]{MNIST 5x50\\ $\epsilon$=0.018}
& 68 | 13 & 19 & 586.5 & 254.9
& \textbf{70 | 13 $_{\uparrow 2}$} & \textbf{17 $_{\downarrow 2}$} & \textbf{537.1 $_{\downarrow 8.4\%}$} & \textbf{199.7 $_{\downarrow 21.7\%}$} \\
\makecell[l]{MNIST 5x80\\ $\epsilon$=0.019}
& 72 | 4 & 24 & 736.1 & 336.8
& \textbf{74 | 4 $_{\uparrow 2}$} & \textbf{22 $_{\downarrow 2}$} & \textbf{704.5 $_{\downarrow 4.3\%}$} & \textbf{301.9 $_{\downarrow 10.4\%}$} \\
\makecell[l]{MNIST 6x100\\ $\epsilon$=0.019}
& 61 | 1 & 38 & 1085.7 & 525.2
& \textbf{63 | 1 $_{\uparrow 2}$} & \textbf{36 $_{\downarrow 2}$} & \textbf{1045.1 $_{\downarrow 3.7\%}$} & \textbf{472.8 $_{\downarrow 10.0\%}$} \\
\makecell[l]{MNIST 9x100\\ $\epsilon$=0.018}
& 52 | 0 & 48 & 1254.8 & 566.6
& \textbf{55 | 0 $_{\uparrow 3}$} & \textbf{45 $_{\downarrow 3}$} & \textbf{1202.3 $_{\downarrow 4.2\%}$} & \textbf{486.6 $_{\downarrow 14.1\%}$} \\
\makecell[l]{MNIST 10x80\\ $\epsilon$=0.015}
& 73 | 0 & 27 & 840.5 & 411.5
& \textbf{73 | 1 $_{\uparrow 1}$} & \textbf{26 $_{\downarrow 1}$} & \textbf{746.8 $_{\downarrow 11.0\%}$} & \textbf{288.9 $_{\downarrow 29.8\%}$} \\
\makecell[l]{MNIST 9x200\\ $\epsilon$=0.018}
& 21 | 0 & 79 & 1790.5 & 1{,}000.6
& \textbf{22 | 0 $_{\uparrow 1}$} & \textbf{78 $_{\downarrow 1}$} & \textbf{1758.2 $_{\downarrow 1.8\%}$} & \textbf{870.1 $_{\downarrow 13.0\%}$} \\
\makecell[l]{VNNCOMP  5x100\\ $\epsilon$=0.019}
& 56 | 1 & 43 & 1148.5 & 505.9
& \textbf{60 | 1 $_{\uparrow 4}$} & \textbf{39 $_{\downarrow 4}$} & \textbf{1077.2 $_{\downarrow 6.2\%}$} & \textbf{393.5 $_{\downarrow 22.2\%}$} \\
\makecell[l]{VNNCOMP  8x100\\ $\epsilon$=0.018}
& 40 | 0 & 60 & 1485.2 & 712.4
& \textbf{46 | 0 $_{\uparrow 6}$} & \textbf{54 $_{\downarrow 6}$} & \textbf{1384.0 $_{\downarrow 6.8\%}$} & \textbf{505.4 $_{\downarrow 29.1\%}$} \\
\makecell[l]{CIFAR10 5x50\\ $\epsilon$=0.010}
& 10 | 73 & 17 & 771.1 & 501.7
& \textbf{10 | 75 $_{\uparrow 2}$} & \textbf{15 $_{\downarrow 2}$} & \textbf{629.6 $_{\downarrow 18.4\%}$} & \textbf{346.4 $_{\downarrow 31.0\%}$} \\
\makecell[l]{CIFAR10 6x80\\ $\epsilon$=0.004}
& 58 | 26 & 16 & 853.4 & 635.0
& \textbf{60 | 27 $_{\uparrow 3}$} & \textbf{13 $_{\downarrow 3}$} & \textbf{652.8 $_{\downarrow 23.5\%}$} & \textbf{416.3 $_{\downarrow 34.4\%}$} \\
\bottomrule
\end{tabular}%
}
\endgroup
\end{table}

\begin{figure}[htbp]
  \centering
  \begin{subfigure}[t]{0.495\textwidth}
    \includegraphics[width=\textwidth]{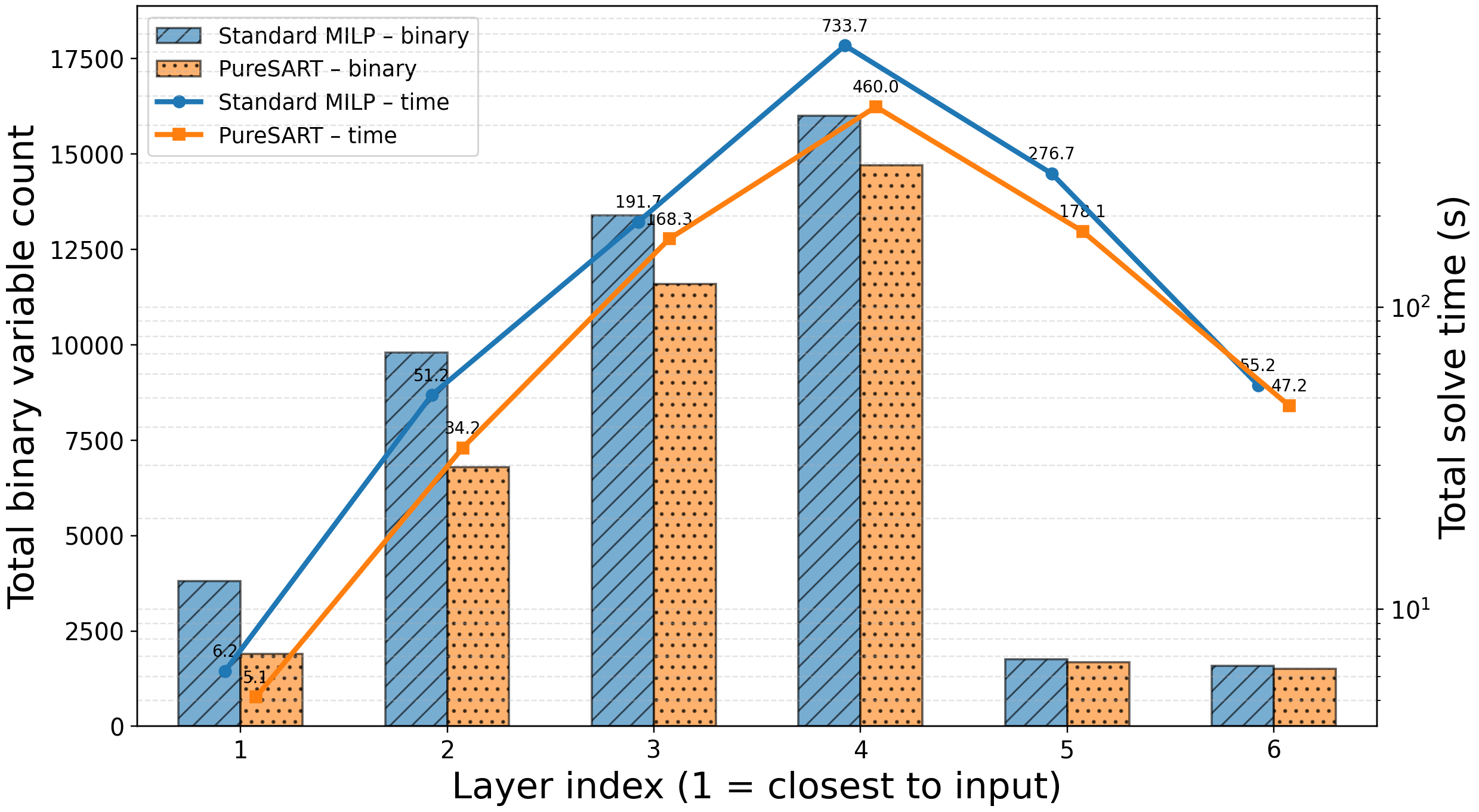}
    \caption{Property ID.18}
  \end{subfigure}
  \begin{subfigure}[t]{0.495\textwidth}
    \includegraphics[width=\textwidth]{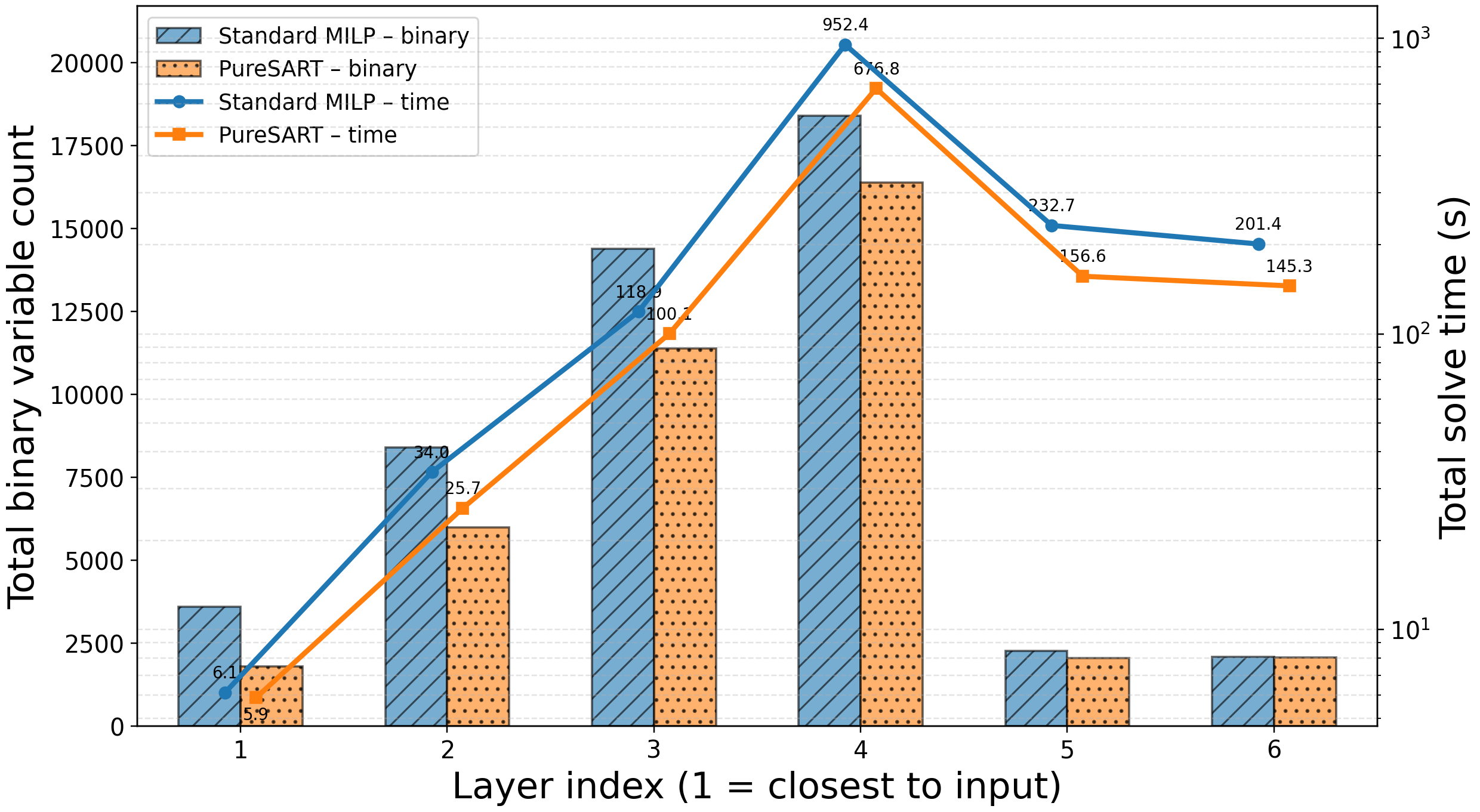}
    \caption{\makebox[0.3\linewidth][l]{Property ID.72}}
  \end{subfigure}
  \caption{\textbf{Layer-wise binary variables and time.}
Standard MILP vs.\ PureSART, ERAN 5$\times$100 (IDs 18, 72).}

  \label{StandardMILPvsPureSART}
\end{figure}

To quantitatively evaluate the effectiveness of the proposed SART encoding itself, we first conducted an independent assessment of the Pure SART encoding. This encoding introduces no abstraction; all constraints are in a complete form and it is compared against the standard MILP encoding. The experimental results are shown in ~\Cref{no_abstraction1}.
The results indicate that the SART encoding significantly improves verification efficiency, achieving more successful verifications and fewer timed-out samples. In statistics including timed-out samples, the SART encoding yields a performance improvement ranging from 1.8\% to 23.5\% compared to the standard encoding. However, limited by timeout truncation, this metric cannot fully reflect the difference in encoding performance.
For a more accurate evaluation of encoding performance, we further excluded samples with timeouts and compared the average verification time over the remaining samples. In this case, the SART encoding achieves a time improvement of 10.0\% to 34.4\% over the standard encoding, surpassing the performance lower bound of the standard MILP encoding.
As shown in \Cref{StandardMILPvsPureSART}, we compare Standard MILP and PureSART on two properties (IDs 18 and 72) from the VNN-COMP ERAN 5$\times$100 model.
For each layer, bars denote the total number of binary variables accumulated when computing bounds for all neurons, and the line reports the corresponding per-layer solve time (right y-axis).
PureSART reduces binary variables on all layers and yields consistent time savings.
For difficult cases that were close to timing out at 2000\,s, after increasing the time limit to 20000\,s, we confirmed that \textsc{SART} achieves more significant improvements on benchmarks with longer runtimes; the results are shown in ~\Cref{tab:near-timeout-hard-cases}.

\begin{table}[t]
\centering
\caption{$\epsilon=0.019$, VNN-COMP 5$\times$100 near-timeout hard cases; runtime (s) with timeout=20000\,s}
\setlength{\tabcolsep}{2.8pt}
\small
\begin{adjustbox}{max width=\linewidth}
\begin{tabular}{ccccccccc}
\toprule
Method (No-Abstraction) & ID.4 & ID.6 & ID.5 & ID.42 & ID.44 & ID.46 & ID.83 \\
\midrule
 \textbf{Our SART} & \textbf{3500.54} & \textbf{5865.26} & \textbf{11219.94} & \textbf{3303.81} & \textbf{1723.20} & \textbf{4790.73} & \textbf{3376.61} \\
Standard-MILP & 4894.93 & 15167.95 & 14118.97 & 6985.35 & 3253.16 & 6834.74 & 6149.84 \\
\bottomrule
\end{tabular}
\end{adjustbox}
\label{tab:near-timeout-hard-cases}
\end{table}

\subsection{Performance of LayerABS}

\begin{table}[htbp]
    \centering
    \small
    \caption{Verification Results on Different Models}
    \label{tab:verification}
    \resizebox{0.75 \textwidth}{!}{%
    \begin{tabular}{p{2.0cm} p{2.0cm} c c c c c c}
        \toprule
        Method & Metric & 5x50 & 5x80 & 6x100 & 9x100 & 10x80 & 9x200 \\
        \midrule
        & $\epsilon$& 0.018 & 0.019 & 0.019 & 0.018 & 0.015 & 0.018 \\ 
        \midrule
        \multirow{3}{*}{\textbf{LayerABS}} 
        & safe | unsafe & \textbf{75 | 12} & \textbf{78 | 5} & \textbf{82 | 1} & \textbf{84} & \textbf{89 | 1} & \textbf{75} \\
        & timeout & \textbf{13} & \textbf{17} & \textbf{17} & \textbf{16} & \textbf{10} & \textbf{25} \\
        & Time(s) & \textbf{396.2} & \textbf{449} & \textbf{440.3} & \textbf{458.5} & \textbf{281.5} & \textbf{776.6} \\
        \midrule
        \multirow{3}{*}{MIPVerify} 
        & safe | unsafe & 69 | 11 & 68 | 2 & 65 | 1 & 67 & 77 & 53 \\
        & timeout & 20 & 30 & 34 & 33 & 23 & 47 \\
        & Time(s) & 599.8 & 807.2 & 925 & 964.2 & 674.5 & 1199.4 \\
        \midrule
        \multirow{3}{*}{Marabou} 
        & safe | unsafe & 39 | 13 & 37 | 9 & 38 | 4 & 30 | 1 & 45 & 38 | 2 \\
        & timeout & 48 & 54 & 58 & 69 & 55 & 60 \\
        & Time(s) & 1003.9 & 1158.6 & 1213.6 & 1385 & 1103.1 & 1202.5 \\
        \midrule
        \multirow{3}{*}{\makecell{$\beta$-CROWN \\ BaB (CPU)}} 
        & safe | unsafe & 48 & 48 & 51 & 45 & 58 & 59 \\
        & timeout & 52 & 52 & 49 & 55 & 42 & 41 \\
        & Time(s) & 1036.8 & 1049.4 & 992 & 1110.6 & 843.9 & 828.4 \\
        \midrule
        \multirow{3}{*}{\makecell{$\beta$-CROWN \\ BaB (GPU)}} 
        & safe | unsafe & 47 & 49 & 51 & 45 & 58 & 59 \\
        & timeout & 53 & 51 & 49 & 55 & 42 & 41 \\
        & Time(s) & 1047.1 & 1045.5 & 989.9 & 1111 & 844.6 & 829.9 \\
        \midrule
        \multirow{3}{*}{DeepPoly} 
        & safe & 14 & 19 & 32 & 28 & 43 & 38 \\
        & unknown & 86 & 81 & 68 & 72 & 57 & 62 \\
        & Time(s) & 1 & 1.7 & 2.7 & 5.1 & 4 & 14.2 \\
        \midrule
        \multirow{3}{*}{$\alpha$-CROWN} 
        & safe & 24 & 30 & 41 & 35 & 50 & 47 \\
        & unknown & 76 & 70 & 59 & 65 & 50 & 53 \\
        & Time(s) & 0.5 & 0.6 & 0.9 & 1.9 & 2.2 & 3.6 \\
        \bottomrule
    \end{tabular}%
    }
\end{table}

We evaluated the overall performance of LayerABS, which integrates the newly proposed SART encoding with DeepPoly-based abstract techniques. Experimental results are shown in Table~\ref{tab:verification}. Compared to state-of-the-art complete verifiers, our proposed layer-wise progressive tightening complete verification algorithm significantly reduces verification time while achieving higher verification accuracy.
Compared to MIPVerify, our method reduces the timeout rate by 7\%--22\% and shortens verification time by 33.9\%--58.3\%. A key challenge in complete verification lies in proving optimality: although heuristic methods can quickly find optimal solutions, proving that their duality gap is zero often incurs high cost. Compared with MIPVerify, our algorithm models unstable neurons more efficiently by avoiding introducing independent binary variables for all unstable neurons, which significantly alleviates the complexity of MIP solving.
As for Marabou, it introduces DeepPoly as a preprocessing step based on the original Reluplex. Although this preprocessing improves efficiency, instances not verified by DeepPoly still cause Marabou to suffer from a high timeout rate, which is also reflected in the accuracy difference between Marabou and DeepPoly. In contrast, our method reduces its timeout rate by 35\%--53\% and shortens verification time by 35.4\%--74.5\%.
The $\beta$-CROWN BaB first employs $\alpha$-CROWN, which has better precision than DeepPoly, for abstract verification, and further combines it with a pure branch-and-bound (BaB) strategy to achieve completeness. This method requires precise branching for each unstable neuron, resulting in high timeout rates on both CPU and GPU. Our method reduces the timeout rate by 16\%--40\% and verification time by 6.3\%--66.7\% based on this. This indicates that although BaB can independently complete complete verification, it still cannot avoid the branch explosion problem caused by unstable neurons; its fundamental difficulty is similar to that of MIPVerify.
It is worth emphasizing that as a fundamental improvement in MIP modeling of unstable neurons, our method exhibits good complementarity with existing complete verifiers and can be integrated as a general-purpose building block into other verification frameworks to further enhance their modeling capability and solving efficiency.
\begin{figure}[htbp]
    \centering
    \includegraphics[width=0.7\linewidth]{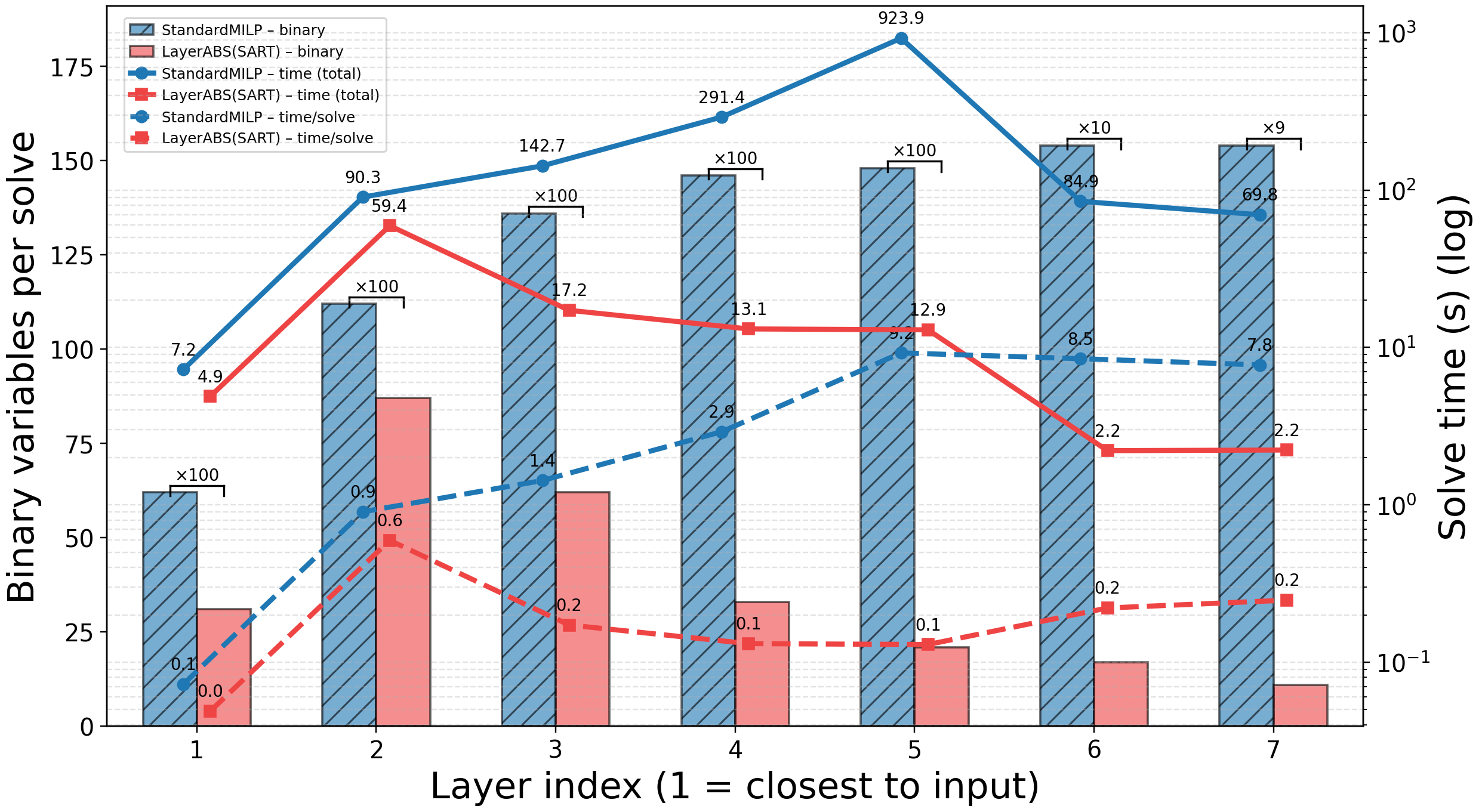}
    \caption{Standard MILP vs.\ LayerABS (SART): binary variables and  time on MNIST 6$\times$100 (property 50).}
\label{fig:layerwise-layerabs}
\end{figure}
As shown in \Cref{fig:layerwise-layerabs}, we compare Standard MILP with LayerABS (complete constraints encoded using SART) on MNIST 6$\times$100, property ID~50.
For each layer, bars report the per-solve binary-variable count (the cumulative constraints needed to solve one neuron).
The ``$\times N$'' label gives the number of neurons in that layer (for example, $\times100$ means 100 neurons).
The solid line shows the total time to process all neurons in the layer (right y-axis), and the dashed line shows the time per solve.
LayerABS dramatically reduces the per-solve binary-variable count on every layer and provides large speedups in deeper layers, often by an order of magnitude.


\noindent \textbf{Fallback Frequency.}
Since LayerABS consists of three cascaded verification strategies with different granularities(Stage1--Stage3), we present in ~\Cref{strategy_freq} the frequency at which it falls back to higher computational cost strategies to comprehensively reflect its solving capability. 
``-'' indicates exceeding the time limit of 2000 seconds.

\begin{table}[htbp]
\centering
\caption{Verification Strategy Fallback Frequency (\%)}
\label{strategy_freq}
\resizebox{0.96\textwidth}{!}{%
\begin{tabular}{@{}cccccccc@{}}
\toprule
Strategy                       & Metric                  & 5x50                   & 5x80                        & 6x100                    & 9x100               & 10x80                        & 9x200            \\ \midrule
\multirow{2}{*}{Stage 1}      & safe                    & 14                     & 19                          & 32                       & 28                  & 43                           & 38               \\
                               & Time(s)                 & 0.8                    & 1.3                         & 2.3                      & 3.5                 & 2.9                          & 10.3             \\ \midrule
\multirow{2}{*}{Stage 2}       & safe | timeout          & 47 | 5                 & 51 | 9                      & 49 | 10                  & 54 | 8              & 45 | 7                       & 37 | 25          \\
                               & Time(s)                 & 26.9 | -          & 101.2  |  -            & 189.6  |  -        & 190.9  |  -    & 118.7 | -               & 723.9  |  - \\ \midrule
\multirow{2}{*}{Stage 3} & safe | unsafe | timeout & 14 | 12 | 8            & 8 | 8 | 5                   & 1 | 1 | 7                & 2 | 0 | 8           & 1 | 1 | 3                    & 0 | 0 | 0         \\
                               & Time(s)                 & 546.4 | 391.0 | - & 320.5  |  625.9    | - & 152.7 |  472.8  | - & 1696.3 | 0 | - & 1415.4  | 1234.1 |   - & 0 | 0 | 0        \\ \bottomrule
\end{tabular}%
}
\end{table}

\subsection{Performance of Incomplete-LayerABS}

\begin{table}[htbp]
\centering
\caption{Verification on MNIST: PRIMA vs.\ Incomplete-LayerABS. Time(s) = mean runtime.}
\label{tab:prima-vs-layerabs}
\begingroup
\setlength{\tabcolsep}{3pt}      
\renewcommand{\arraystretch}{0.95} 
\footnotesize
\resizebox{0.6\textwidth}{!}{%
\begin{tabular}{@{}lcccccc@{}}
\toprule
\multirow{2}{*}{Model}
& \multicolumn{3}{c}{PRIMA (SOTA multi-neuron)}
& \multicolumn{3}{c}{\textbf{Incomplete-LayerABS}} \\
\cmidrule(lr){2-4} \cmidrule(lr){5-7}
& safe $\uparrow$ & unknown $\downarrow$ & Time(s) $\downarrow$
& \textbf{safe $\uparrow$} & \textbf{unknown $\downarrow$} & \textbf{Time(s) $\downarrow$} \\
\midrule
\makecell[l]{MNIST 5x50\\ $\epsilon$=0.018}
& 71 & 29 & 113.0
& \textbf{74 $_{\uparrow 3}$} & \textbf{26 $_{\downarrow 3}$} & \textbf{75.5 $_{\downarrow 33.2\%}$} \\
\makecell[l]{MNIST 5x80\\ $\epsilon$=0.019}
& 73 & 27 & 213.0
& \textbf{78 $_{\uparrow 5}$} & \textbf{22 $_{\downarrow 5}$} & \textbf{144.3 $_{\downarrow 32.2\%}$} \\
\makecell[l]{MNIST 6x100\\ $\epsilon$=0.019}
& 80 & 20 & 330.9
& \textbf{84 $_{\uparrow 4}$} & \textbf{16 $_{\downarrow 4}$} & \textbf{225.7 $_{\downarrow 31.8\%}$} \\
\makecell[l]{MNIST 9x100\\ $\epsilon$=0.018}
& 84 & 16 & 395.0
& \textbf{85 $_{\uparrow 1}$} & \textbf{15 $_{\downarrow 1}$} & \textbf{238.8 $_{\downarrow 39.5\%}$} \\
\makecell[l]{MNIST 10x80\\ $\epsilon$=0.015}
& 90 & 10 & 243.5
& \textbf{91 $_{\uparrow 1}$} & \textbf{9 $_{\downarrow 1}$} & \textbf{134.9 $_{\downarrow 44.6\%}$} \\

\makecell[l]{MNIST 9x200\\ $\epsilon$=0.018}
& 79 & - & 885.5
& \textbf{86 $_{\uparrow 7}$} & - & \textbf{602.7 $_{\downarrow 31.9\%}$} \\

\bottomrule
\end{tabular}%
}

\begin{minipage}{0.6\textwidth}
\tiny \emph{Note. For MNIST \(9\times 200\), some runs timed out; since timeouts were neither separated from nor counted as \textit{unknown}, the \textit{unknown} entry is left blank.}
\end{minipage}

\endgroup
\end{table}

We focus on the core challenge of unstable-neuron encoding, which arises in both complete and incomplete verification. Removing Stage~3 from LayerABS yields Incomplete-LayerABS, which is an incomplete method.
\Cref{tab:prima-vs-layerabs} compares PRIMA with Incomplete-LayerABS on MNIST under identical abstraction: both methods use DeepPoly, and the MIP solver time limit is set to 30s for fairness. For completeness experiments reported elsewhere, we do not impose a time limit on LayerABS in order to preserve completeness.
Under this setting, Incomplete-LayerABS achieves more \textit{safe} results and reduces the average verification time by 31.8\%-44.6\% relative to PRIMA. These results indicate that the SART-based progressive tightening effectively shrinks the search space and improves the efficiency of  verification.

\subsection{Ablation Study on the SART Encoding in LayerABS}

\begin{table}[htbp]
\centering
\caption{Performance of MILP and SART Encoding in LayerABS. 
Time (w/ to) = mean including timeouts at the time limit; 
Time (excl. to \& DeepPoly) = mean over MILP-only cases (excluding timeouts and DeepPoly-solved).}
\label{tab:verification_results}
\begingroup
\setlength{\tabcolsep}{5pt}
\renewcommand{\arraystretch}{1.2}
\resizebox{\textwidth}{!}{%
\begin{tabular}{@{}lcccccccc@{}}
\toprule
\multirow{2}{*}{Model}
& \multicolumn{4}{c}{LayerABS (Standard MILP Encoding)}
& \multicolumn{4}{c}{\textbf{LayerABS (Our SART Encoding)}} \\
\cmidrule(lr){2-5} \cmidrule(lr){6-9}
& Safe | Unsafe (DP) $_{sum\uparrow}$ & Timeout $_{\downarrow}$ & Time (w/ to) $_{\downarrow}$ & Time (excl. to \& DP) $_{\downarrow}$
& \textbf{Safe | Unsafe (DP) $_{sum\uparrow}$} & \textbf{Timeout $_{\downarrow}$} & \textbf{Time (w/ to) $_{\downarrow}$} & \textbf{Time (excl. to \& DP) $_{\downarrow}$} \\
\midrule
\makecell[l]{MNIST 5x50\\ $\epsilon$=0.018}
& 73 | 12 (14) & 15 & 424.3 & 174.8
& \textbf{75 | 12 (14) $_{\uparrow 2}$} & \textbf{13 $_{\downarrow 2}$} & \textbf{396.2 $_{\downarrow 6.6\%}$} & \textbf{141.6 $_{\downarrow 19.0\%}$} \\
\makecell[l]{MNIST 5x80\\ $\epsilon$=0.019}
& 77 | 5 (19) & 18 & 491.0 & 207.1
& \textbf{78 | 5 (19) $_{\uparrow 1}$} & \textbf{17 $_{\downarrow 1}$} & \textbf{449.0 $_{\downarrow 8.6\%}$} & \textbf{150.6 $_{\downarrow 27.3\%}$} \\
\makecell[l]{MNIST 6x100\\ $\epsilon$=0.019}
& 82 | 1 (32) & 17 & 465.2 & 243.3
& \textbf{82 | 1 (32)} & \textbf{17} & \textbf{440.3 $_{\downarrow 5.4\%}$} & \textbf{194.4 $_{\downarrow 20.1\%}$} \\
\makecell[l]{MNIST 9x100\\ $\epsilon$=0.018}
& 83 | 0 (28) & 17 & 471.6 & 236.2
& \textbf{84 | 0 (28) $_{\uparrow 1}$} & \textbf{16 $_{\downarrow 1}$} & \textbf{458.5 $_{\downarrow 2.8\%}$} & \textbf{219.9 $_{\downarrow 6.9\%}$} \\
\makecell[l]{MNIST 10x80\\ $\epsilon$=0.015}
& 88 | 0 (43) & 12 & 299.8 & 129.1
& \textbf{89 | 1 (43) $_{\uparrow 2}$} & \textbf{10 $_{\downarrow 2}$} & \textbf{281.5 $_{\downarrow 6.1\%}$} & \textbf{118.7 $_{\downarrow 8.1\%}$} \\
\makecell[l]{MNIST 9x200\\ $\epsilon$=0.018}
& 74 | 0 (38) & 26 & 812.1 & 785.9
& \textbf{75 | 0 (38) $_{\uparrow 1}$} & \textbf{25 $_{\downarrow 1}$} & \textbf{776.6 $_{\downarrow 4.4\%}$} & \textbf{699.2 $_{\downarrow 11.0\%}$} \\
\bottomrule
\end{tabular}%
}
\endgroup
\end{table}

\begin{figure}[htbp]
    \centering
    \includegraphics[width=0.7\linewidth]{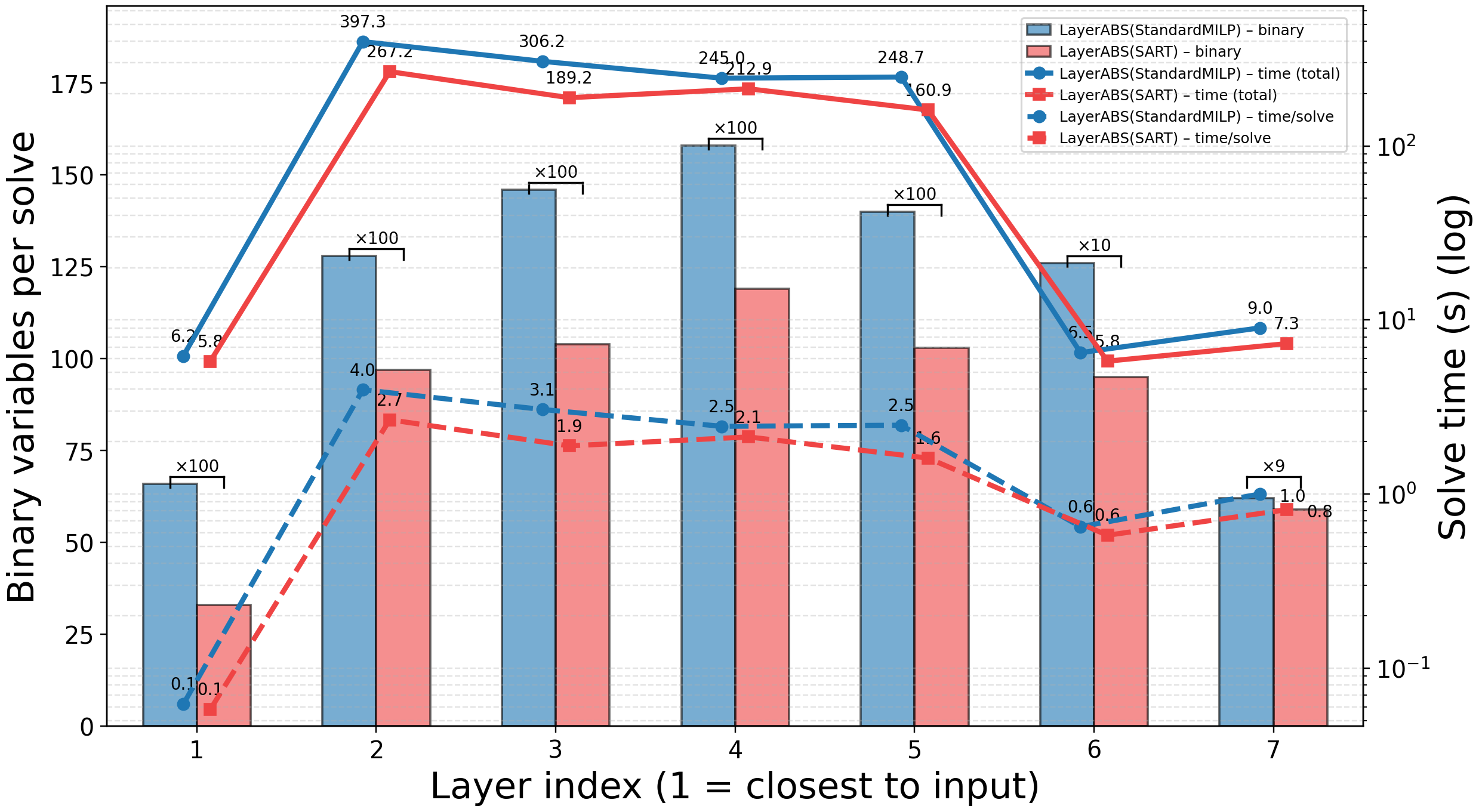}
\caption{LayerABS(MILP) vs.\ LayerABS(SART): binary variables and time on MNIST 6$\times$100 (property 89).}
\label{fig:layerwise-layerabs-sart-vs-milp}
\end{figure}

The performance improvement of LayerABS originates from the newly introduced SART encoding and the abstraction technique. Since the abstraction strategy is known to enhance performance, it is necessary to distinguish the source of acceleration. Section~\ref{pure} has independently evaluated the SART encoding, and this section compares the performance between the standard MILP and the SART encoding within LayerABS. The results ( \Cref{tab:verification_results}) indicate that the SART encoding improves efficiency, achieving an increase of 2.8\%--8.6\% when including samples with timeouts. However, the truncation due to timeout and the abstraction technique obscure the performance difference, failing to fully reflect the advantage of the encoding. After excluding samples with timeouts and those successfully verified by DeepPoly, the efficiency gain expands to 6.9\%--27.3\%, fully validating the effectiveness of the SART encoding in LayerABS.
As shown in \Cref{fig:layerwise-layerabs-sart-vs-milp}, we compare LayerABS(MILP) and LayerABS(SART), which differ only in the complete-constraint encoding (MILP vs.\ SART), on MNIST 6$\times$100, property ID~89.
For each layer, bars report the per-solve binary-variable count (the cumulative constraints needed to solve one neuron).
The ``$\times N$'' label gives the number of neurons in that layer.
The solid line shows the total time to process all neurons in the layer (right y-axis), and the dashed line shows the time per solve.
Even with abstraction enabled, the SART variant reduces the per-solve binary-variable count at every layer and yields lower per-layer times.

\section{Related Work}
Complete methods (MILP/SMT/BaB~\cite{anderson2020strong,tjeng2017evaluating,katz2017reluplex,ehlers2017formal,xu2020reluplex,wang2021beta,shi2024neural}) guarantee termination but scale poorly. Incomplete methods rely on interval/symbolic~\cite{wang2018formal,wang2018efficient}, linear/abstract (DeepPoly, CROWN/$\alpha$-CROWN, Fast-Lin/Lip~\cite{singh2019abstract,zhang2018efficient,xu2020fast,weng2018towards}) with known single-neuron gaps~\cite{salman2019convex}, and multi-neuron relaxations (kPoly, OptC2V, PRIMA, Merging~\cite{singh2019beyond,tjandraatmadja,muller2022prima,elboher2020abstraction}). SDP/dual relaxations~\cite{raghunathan2018semidefinite,batten2021efficient} improve tightness with efficient solvers~\cite{dathathri2020enabling}. We instead reformulate the \emph{encoding} (SART), orthogonal to these lines.
Although weight sign decomposition has been explored in incomplete verification, its practical use has been limited by efficiency concerns, for example, IBP~\cite{dvijotham2018training} deems it computationally sub-optimal for interval propagation and thus avoids it, or by high computational overhead, for instance, the method proposed by Elboher \emph{et al.} \cite{elboher2020abstraction} can inflate the network size by about $4\times$. In this paper, we revisit its value for complete MILP-based verification: when weight sign separation is combined with an absolute-value transformation, the sign distribution of weights directly determines the number of binary variables required by the MILP encoding. Leveraging this newly identified mathematical dependency, we precisely quantify and exploit it by redefining the optimization problem, thereby substantially reducing the number of required binary variables.

\section{Discussion on Generality}

We summarize the generality of the encoding and present three complementary axes: activations, norm bounded perturbations and regions, and integration with other solvers.

\begin{enumerate}[label=\textbf{\arabic*.}, leftmargin=*]

\item \textbf{Activation functions.}
Although our development centers on ReLU, the encoding extends directly to any piecewise linear activation that admits an absolute value representation \( f(x) = p\,x + q\,\lvert x\rvert + r \) with fixed coefficients.
Leaky ReLU, PReLU, and RReLU all fit this template, so the constraints are instantiated by substituting the corresponding coefficients (for example, \( p = \tfrac{1+\alpha}{2} \), \( q = \tfrac{1-\alpha}{2} \), and \( r = 0 \)).
For the standard smooth Tanh, SART does not provide complete support if we require algebraic exactness, since mixed arithmetic involving transcendental functions is generally undecidable , as indicated by Richardson’s theorem~\cite{richardson1969some}. In practice, the impact is limited because lightweight models often use Hard-Tanh as an efficient substitute, and Hard-Tanh can be exactly expressed as $0.5(\lvert x+1\rvert-\lvert x-1\rvert)$, making it amenable to exact SART verification.

\item \textbf{Norm bounded perturbations and regions.}
Vector norms are used only to define the input perturbation set. The encoding of unstable ReLU neurons does not depend on the chosen norm. The method therefore applies to \(\ell_\infty\), \(\ell_1\), and \(\ell_2\). More broadly, if the outer verifier can describe an input or output region, the encoding works with it. Common cases include polyhedral sets and norm balls. These choices do not change the core encoding.

\item \textbf{Integration with other solvers.}
The encoding is a low level, theory driven scheme that operates at the constraint level. It can be instantiated within complete and incomplete verification pipelines, including mixed integer linear programming workflows, branch and bound workflows, and relaxation based workflows. Representative systems include Marabou, MIPVerify, $\beta$-CROWN BaB, and PRIMA, where the encoding can replace or strengthen the treatment of unstable ReLUs.

\end{enumerate}







\section{Conclusion}

This work introduces the Sign--Absolute Reformulation Theory (SART), which breaks the rigid one-to-one correspondence between unstable ReLU neurons and binary variables via weight-sign decomposition and an absolute-value equivalence of ReLU. We derive linearization criteria for non-nested absolute values, identify when absolute-value terms in the objective or constraints govern problem scale, and establish conditions under which the number of binaries per unstable neuron can be reduced. Under interval-complete bound computation, we prove a tight limit: the final layer admits a 50\% reduction in binary variables, whereas intermediate layers are non-reducible under the assumption \(P \neq NP\).
Crucially, we show that this apparent ``last-layer-only'' gain can be lifted to the whole network by recasting verification as a sequential process. Our verifier, LayerABS, implements SART-based progressive tightening that treats intermediate layers as temporary final layers. Each step realizes the optimal 50\% local reduction at the active tail layer while producing high-precision bounds that propagate across the model, fix many activations early, shrink the search space, and thereby reduce the number of binaries subsequently introduced. In effect, final-layer compression becomes a network-wide acceleration mechanism.
We reveal a link between weight-sign symmetry and complexity: when unstable neurons have near-symmetric signs (positive/negative counts equal or differ by at most one), worst-case verification complexity attains the theoretical optimum. 
Finally, SART is validated as a general-purpose encoding: we evaluate abstraction-free PureSART, and integrate SART with abstractions to build the complete LayerABS and its incomplete variant Incomplete-LayerABS. Across benchmarks, these SART-based instances surpass strong baselines, showing that principled encoding, paired with progressive tightening, turns theoretical limits into end-to-end verification speedups.

\section*{Acknowledgments}
This work was supported by the National Natural Science Foundation of China (No. 62472316), the Shanghai 2023 “Science and Technology Innovation Action Plan” Special Project for Key Technical Breakthrough of Blockchain (No. 23511100800), and the National Natural Science Foundation of China (No. 62406227).

\section*{Data Availability Statement}
The source code and benchmarks used in this study, along with supplementary materials, are publicly available at
\url{https://github.com/babutiger/SART}.

\bibliographystyle{ACM-Reference-Format}
\bibliography{reference.bib}

\end{document}